%% file: main.tex
\documentclass[msom,nonblindrev]{informs3}

\OneAndAHalfSpacedXII % Current default line spacing
\EquationsNumberedThrough    % Default: (1), (2), ...
\usepackage{natbib}
\bibpunct[, ]{(}{)}{,}{a}{}{,}%
\def\bibfont{\small}%
\input{preamble}

\newcommand{\pub}{\mathtt{pub}} % \pub(\type) is public type

\newcommand{\type}{\scr{X}}
\newcommand{\types}{\cal{X}}
\newcommand{\pubtype}{\type^{\mathtt{\pub}}}
\newcommand{\pubtypes}{\types^{\mathtt{\pub}}}
\newcommand{\msg}{M}
\newcommand{\msgs}{\cal{M}}
\newcommand{\arm}{\cal{A}}
\newcommand{\arms}{\cal{K}}
\newcommand{\aux}{\mathtt{aux}} % auxiliary feedback
\newcommand{\AUX}{\mathtt{AUX}} % feedback set

\newcommand{\rew}{y}
\newcommand{\rewE}{\E_{\mathtt{rew}}} % expectation over rewards only
\newcommand{\outD}{D^{\mathtt{out}}} % outcome distribution
\newcommand{\rewD}{D^{\mathtt{rew}}} % outcome distribution

\newcommand{\model}{u}
\newcommand{\trueModel}{u^*}
\newcommand{\MODELS}{\cal{U}}
\newcommand{\prior}{\cal{P}_0} % prior over models
\newcommand{\pimod}{\pi_{\mathtt{mod}}} % model-selection policy

\newcommand{\menu}{\mathtt{menu}} % recommendation menu

\newcommand{\smap}{\scr{Q}} % semantic map
\newcommand{\epsTS}[1][(\smap)]{\mathtt{\varepsilon}_0{#1}}
\newcommand{\delTS}[1][(\smap)]{\mathtt{\delta}_{0}{#1}}
\newcommand{\NTS}[1][(\smap)]{N_{0}{#1}}

\newcommand{\ourAlg}{filtered posterior sampling\xspace}
\newcommand{\OurAlg}{Filtered posterior sampling\xspace}

\newcommand{\ols}{filtered least squares\xspace}
\newcommand{\OLS}{Filtered least squares\xspace}

\newcommand{\UCB}{UCB/Greedy\xspace}

\newcommand{\warmup}{\mathtt{WarmUp}} % warm-up data

\newcommand{\emin}[1][{[T_0]}]{\lambda_{{#1}}}
\newcommand{\eminprime}[1][{[T_0]}]{\lambda^{\mathtt{had}}_{{#1}}}

\newtheorem{theorem}{Theorem}
\newtheorem{lemma}{Lemma}
\newtheorem{corollary}{Corollary}
\newtheorem{definition}{Definition}
\newtheorem{assumption}{Assumption}
\newtheorem{remark}{Remark}[section]
\newtheorem{fact}{Fact}

\TITLE{Incentivized Exploration via Filtered Posterior Sampling}

\ARTICLEAUTHORS{%
\AUTHOR{Anand Kalvit}
\AFF{Stanford University, \EMAIL{akalvit@stanford.edu}} %, \URL{}}
\AUTHOR{Aleksandrs Slivkins}
\AFF{Microsoft Research, \EMAIL{slivkins@microsoft.com}}
\AUTHOR{Yonatan Gur}
\AFF{Stanford University, Netflix, \EMAIL{ygur@\{stanford.edu, netflix.com\}}}
} % end of the block

\ABSTRACT{

\input{abstract}

}

\begin{document}

\maketitle

% Paper body

\input{introduction}

\subsection{Related work}
\label{sec:related}
\input{sec-related}

\input{model-and-prelim}

\section{\OurAlg: Main Results and Corollaries}
\label{section:main-TS}
\input{main-result-section}

%\section{Implications of Theorem~\ref{thm:general}}
%\label{section:implications}
\input{implications}

\section{\OurAlg: Additional Corollaries}
\label{sec:TS-other}
\input{sec-other}

\input{IE-via-others}

\bibliographystyle{informs2014} 
\bibliography{bib-abbrv,bib-slivkins,bib-AGT,bib-bandits,bib-anand}

\begin{appendices}

\input{standard_definitions}

\input{Main_proof}

\input{OLS_proof}
\input{UCB_proof}
\input{corollary_proofs}
\input{martingale_stuff}

\end{appendices}

\end{document}

%% file: preamble.tex
\usepackage{hyperref} % must be loaded before cleveref

\usepackage{color-edits}
\addauthor{as}{red}      % as for Alex
\addauthor{ak}{blue}     % ak for Anand
\addauthor{yg}{magenta}  % yg for Yoni
\usepackage[title,titletoc]{appendix}
\let\theoremstyle\relax

\usepackage{amsthm}
\usepackage{amssymb}
\usepackage{amsmath}
\usepackage{amsfonts}
\usepackage{graphicx}
\usepackage{color}
\usepackage{bbm}
\usepackage{physics}
\usepackage[mathscr]{euscript}
\usepackage{bm}
\usepackage{verbatim}

\newcommand{\R}{\mathbb{R}}
\newcommand{\N}{\mathbb{N}}
\newcommand{\Prob}[2][]{\mathbb{P}_{#1}\sbr{#2}}
\newcommand{\e}{\mathbf{e}}
\newcommand{\bb}[1]{\mathbb{#1}}
\newcommand{\scr}[1]{\mathscr{#1}}
\newcommand{\ith}[1]{{#1}^{\text{th}}}
\newcommand{\vol}[1]{{\texttt{Vol}_d\left({#1}\right)}}
\newcommand{\pbrac}[1]{{{\left( {#1} \right)}}}
\newcommand{\cbrac}[1]{{\left\lbrace {#1} \right\rbrace}}
\newcommand{\sbrac}[1]{{\left[ {#1} \right]}}
\newcommand{\indi}[1]{{\mathbbm{1}{\cbrac{{#1}}}}}
\newcommand{\epsUCB}{\mathtt{\varepsilon}_{\texttt{UCB}}}

\newcommand{\NUCB}{N_{\texttt{UCB}}}
\newcommand{\subG}{\mathtt{subG}}
\newcommand{\s}{\frak{s}}
\newcommand{\xpub}{x^{\texttt{pub}}}
\newcommand{\xsingleton}{x_0}
\newcommand{\armprime}{\arm^\prime}

\renewcommand{\mod}[1]{\left\lvert{#1}\right\rvert}
\renewcommand{\geq}{\geqslant}
\renewcommand{\leq}{\leqslant}
\renewcommand{\cal}{\mathcal}
\renewcommand{\frak}{\mathfrak}
\newcommand{\E}{\mathbb{E}}

\usepackage{xspace,nicefrac}

\newcommand{\xhdr}[1]{\vspace{1mm} \noindent{\bf #1}}

\newcommand{\ie}{{\em i.e.,~\xspace}}

\newcommand{\eg}{{\em e.g.,~\xspace}}

\newcommand{\rbr}[1]{\left(\,#1\,\right)}
\newcommand{\sbr}[1]{\left[\,#1\,\right]}
\newcommand{\cbr}[1]{\left\{\,#1\,\right\}}

\newcommand{\LDOTS}{\, ,\ \ldots\ ,}
\newcommand{\eps}{\varepsilon}
\newcommand{\refeq}[1]{Eq.~(\ref{#1})}

%% file: abstract.tex
We study \emph{incentivized exploration} (IE) in social learning problems where the principal (a recommendation algorithm) can leverage information asymmetry to incentivize sequentially-arriving agents to take exploratory actions. We identify posterior sampling, an algorithmic approach that is well known in the multi-armed bandits literature, as a general-purpose solution for IE. In particular, we expand the existing scope of IE in several practically-relevant dimensions, from private agent types to informative recommendations to correlated Bayesian priors. We obtain a general analysis of posterior sampling in IE which allows us to subsume these extended settings as corollaries, while also recovering existing results as special cases.

%% file: introduction.tex
\section{Introduction}

%\textbf{High-level problem description.}
A principal (social planner) interacts sequentially with a flow of self-interested agents that each take actions, consume information, and produce information over time. The planner's goal is to maximize the aggregate utility of all agents it interacts with, which necessitates agents to occasionally take exploratory actions that might otherwise be deemed inferior from an empirical standpoint. While such exploratory actions are the cornerstone of online learning as they help the principal learn the best actions over time, they also represent misaligned incentives between the principal and individual agents.
 %However, ; indeed, what's best for one in a given round need not be the best for all across all rounds and vice-versa.
How can a welfare-maximizing principal achieve her goal in the presence of such misaligned incentives? This is the essence of the incentivized exploration problem.

While a desirable alignment of incentives may be achieved via monetary payments to the agents, this often is infeasible, impractical, or unethical. An alternative approach is to leverage \textit{information asymmetry} and use signals (or messages) to incentivize agents to take exploratory actions.
% that benefit the principal (for the sake of maximizing social welfare).
Online learning algorithms are a natural vehicle for studying this problem.

%\medskip
%\subsection{Motivation}
%\akcomment{Discuss immigration, DocFind examples to motivate rankings (as a segue towards general message spaces beyond direct recommendations) ..TBD..}
%\noindent \ygedit{\textbf{Motivation from recommender systems.}

A key motivation for studying incentivized exploration is recommendation systems, broadly construed as platforms that collect and aggregate consumers' signals such as reviews and rankings and present them to subsequent consumers. Such platforms are prevalent in a variety of application domains (including streaming and media services, retail, and transportation, to name a few). Nevertheless, recommendation systems are characterized by three distinctive features that so far have not been adequately treated in prior work on incentivized exploration:
\begin{description}
\item[Private agent types.]
While most prior work focuses on homogeneous consumers, in reality they are heterogeneous, with individual preferences that are not fully observable.

\item[Informative recommendations.]
While most prior work focuses on direct recommendations (and nothing but), in reality consumers prefer to see detailed supporting information.

\item[Correlated priors.]
While most prior work relies on priors/beliefs that are independent across the primitives of the problem, in reality priors/beliefs can exhibit complex correlations, encoding the detailed world-view of the consumers.

\end{description}

The general formulation and analysis we develop in the current work allows us to provide a comprehensive and unified treatment of these three directions. In contrast, prior work contains only partial (albeit quite challenging) results along each direction (see Section~\ref{sec:related}).

%\subsection{Main Contributions}

\xhdr{Main contributions.}
Our main results extend \emph{posterior sampling} and present a unified analysis thereof along the three directions outlined above.%
\footnote{Posterior sampling, and particularly the well known Thompson Sampling algorithm, is a standard approach in multi-armed bandits. The approach is as follows: in each round, compute a posterior distribution over models, sample a model from this distribution, and choose the best arm according to this model. This approach is empirically efficient and theoretically near-optimal in many scenarios. It has been shown to work for incentivized exploration \citep{Selke-PoIE-ec21,CombiIE-neurips22,sellke2023incentivizing}, but only for public agent types, with direct recommendations, and under independence assumptions.}
We also recover existing results on posterior sampling as special cases. Much of our analyses is unified, with some case-specific last steps.

We prove that posterior sampling is compatible with agents' incentives when provided with  warm-up data of sufficient "quality". Crucially, the sufficient amount/quality of warm-up data is determined by the problem structure (such Bayesian prior and the geometry of feasible models and agent types), and is therefore a constant with respect to the time horizon. The warm-up data can be collected by another algorithm within the framework of incentivized exploration, or by some other means such as monetary payments.

The learning problem in our model corresponds to \emph{linear contextual bandits}, a well-studied variant of   multi-armed bandits where an algorithm observes a \emph{context} in each round -- essentially, a feature vector for each arm -- and the expected reward of each arm is linear in its feature vector. This is a standard model for user heterogeneity, as far as bandits are concerned.

The key modification in the algorithm is a fixed and known \emph{semantic map} from the models produced by the algorithm to the ``messages" seen by the agents. The messages are therefore assigned consistent meaning in terms of the model being learned. In particular, the semantic map can enforce the desired ``template" for informed recommendations. It is also essential to handle private agent types.

On the technical frontier, our analysis leverages natural properties of posterior sampling in conjunction with novel problem geometry-dependent concentration bounds for posterior samples. In fact, our technical apparatus lends itself to the analysis of more general learning algorithms as well. This is demonstrated in \S\ref{section:other-algorithms} where we analyze and propose results for a few canonical examples of online learning algorithms other than posterior sampling.

%This paper extends prior literature on incentivized exploration via posterior sampling algorithms (which has predominantly focused on Thompson Sampling) along several practical dimensions.

\xhdr{Additional results.}
Without incentives, our model reduces to multi-armed bandits. Yet, posterior sampling is the only ``native" bandit algorithm known to work for incentivized exploration (whereas all other approaches are custom-designed). Thus, one wonders if some other native bandit algorithms satisfy incentive compatibility, too. Leveraging techniques from our main analysis allows us to now establish this property for three other native bandit algorithms: OLS-Greedy for linear bandits, and UCB and Frequentist-Greedy for stochastic bandits. Nevertheless, our analyses strongly suggest the superiority of posterior sampling as an approach for incentivized exploration.

\xhdr{Organization of the paper.} 
%The rest of this paper is organized as follows: 
\S\ref{section:model} provides defines the general model of incentivized exploration studied in this paper.
\S\ref{section:main-TS} presents our algorithm and formulates the main results, including the general analysis and the main corollaries.
\S\ref{sec:TS-other} instantiates our analysis to a few other special cases, including the guarantees from prior work.
Much of the analysis is deferred to the appendices. 
%contains our main result on incentivized exploration using posterior sampling and general message spaces (Theorem~\ref{thm:general}). \S\ref{section:implications} instantiates Theorem~\ref{thm:general} in several corollaries that span novel (Corollaries~\ref{corollary:CB-private}, \ref{corollary:sleeping}) as well as canonical test cases of incentivized exploration (Corollaries~\ref{corollary:CB-public}, \ref{corollary:bandit}, \ref{corollary:combinatorial}). In \S\ref{section:other-algorithms}, we provide guarantees for other native bandit algorithms: OLS-Greedy for linear models in Theorem~\ref{thm:BIC-OLS}, and UCB and Frequentist-Greedy for $K$-armed bandits in Theorem~\ref{thm:BIC-UCB}. Proofs and other technical analyses are relegated to later sections.

%% file: sec-related.tex
Incentivized exploration has been introduced in \citep{Kremer-JPE14,Che-13} and studied in a series of subsequent papers. The basic model, corresponding to stochastic bandits, has been treated in \citep{Kremer-JPE14,ICexploration-ec15,Selke-PoIE-ec21}. In particular, \citet{Selke-PoIE-ec21} analyze Thompson Sampling for this problem, with direct recommendations (and nothing but) and under independent priors.
Several papers touch upon the three extensions emphasized in this paper. 
\begin{itemize}
\item \citet{Jieming-multitypes-www19} consider private agent types, albeit only for deterministic rewards and
via an inefficient algorithm which does not trade off exploration and exploitation.
\item \citet{Jieming-unbiased18} consider "informative recommendations" of a particular shape, whereby each agent observes the ``sub-history" for a predetermined (and carefully chosen) subset of previous agents. Besides, the guaranteed regret rate scales very suboptimally with the number of arms, compared to Thompson Sampling.
\item \citet{CombiIE-neurips22,sellke2023incentivizing} further study Thompson Sampling and allow the Bayesian priors to be correlated across arms, but either require independence across the primitives of the problem, or consider a specific (uniform) prior.\footnote{Namely, \citet{CombiIE-neurips22} consider the combinatorial semi-bandits and require independence across the "atoms". \citet{sellke2023incentivizing} studies linear contextual bandits (with public types), under a uniform prior.}
\end{itemize}
Similar, but technically incomparable versions of incentivized exploration have been studied, \eg with time-discounted rewards \citep{Bimpikis-exploration-ms17},
inevitable information leakage \citep{Bahar-ec16,Bahar-ec19},
or creating incentives via money
\citep{Frazier-ec14,Kempe-colt18}.

Incentivized exploration is related to the literature on information design \citep{Kamenica-survey19,BergemannMorris-survey19}: essentially, one round of incentivized exploration is an instance of Bayesian persuasion, a central model in this literature.

Absent incentives, our model reduces to multi-armed bandits and various extensions thereof. Bandit problems received a huge amount of attention over the past few decades, \eg see books
    \citep{CesaBL-book,slivkins-MABbook}.
Stochastic $K$-armed bandits
    \citep{Lai-Robbins-85,bandits-ucb1}
is a canonical and well-understood "basic" version of the problem. Linear contextual bandits is one of the most prominent extensions of bandits, studied since
\citep{Langford-www10,Reyzin-aistats11-linear,Csaba-nips11}.
Combinatorial semi-bandits (studied in \S\ref{sec:combi-bandits}) are also well-studied, \citep[\eg][]{Chen-icml13,Kveton-aistats15,MatroidBandits-uai14}.

%Sleeping bandits \citep{sleeping-colt08}

Thompson Sampling \citep{Thompson-1933} is a well-known bandit algorithm, see  \citet{TS-survey-FTML18} for background. It enjoys Bayesian regret bounds which are optimal in the worst case
\citep{Russo-MathOR-14,bubeck2013prior} and improve for some ``nice" priors \citep{Russo-MathOR-14}. Also, it achieves near-optimal "frequentist" regret bounds
\citep{Shipra-colt12,Kaufmann-alt12,Shipra-aistats13}.
Thompson Sampling has been applied to linear contextual bandits, starting from \citet{Shipra-icml13}, and to combinatorial semi-bandits,  \citep[\eg][]{Wen-icml15,Russo-JMLR-16}.

%There has been a line of work orthogonal to ours on mechanisms that incentivize exploration via payment \citep{frazier2014incentivizing, kannan2017fairness, chen2018incentivizing}. There are several known disadvantages of such payment mechanisms, including potential high costs and ethical concerns \citep{groth2010honorarium}. For a detailed discussion, see \citet{slivkins2017incentivizing}. 

%% file: model-and-prelim.tex
\section{Model: Incentivized Exploration}
\label{section:model}

\xhdr{Game protocol.} \emph{Incentivized exploration} is a game between a principal and sequentially arriving agents which proceeds according to the protocol in Figure~\ref{fig:IE-protocol}. %\vspace{-0.6cm}

\begin{figure}[ht]
\centering
\caption{Protocol: Incentivized Exploration}
\begin{tabular}{|p{\textwidth}|}
\hline\vspace{0.1mm}
\emph{Primitives:} action set $\arms$, time horizon $T$, model set $\MODELS$, feedback set $\AUX$;\\
\hspace{16mm}type set $\types$, public type set $\pubtypes$, mapping $\pub:\types\to\pubtypes$.\\
Nature draws the model $\trueModel\in\MODELS$ according to the Bayesian prior $\prior$,\\
\hspace{10mm} and fixes the type sequence $\vec{\type} = (\type_t\in\types:\,t\in[T])$. \\
Principal announces its algorithm (\emph{messaging policy}) $\pi$ and message set $\msgs$. \\
In each round $t = 1,2 \LDOTS T$:
\begin{enumerate}
  \item Agent $t$ arrives with type $\type_t\in\types$ and public type
        $\pubtype_t = \pub(\type_t)\in\pubtypes$; \newline
      the agent observes $t$ and $\type_t$ (and nothing else);
      the principal observes $\pubtype_t$.
  \item Principal sends a message $\msg_t\in\msgs$ to agent $t$, generated according to policy $\pi$.
  \item Agent $t$ chooses some action $\arm_t\in\arms$, collects a reward of $\rew_t\in \R$, and leaves.
\item Principal observes $\arm_t,\rew_t$ and (possibly) some auxiliary feedback $\aux_t\in\AUX$.\vspace{-0.1cm}\end{enumerate}
\\\hline
\end{tabular}
\label{fig:IE-protocol}
\end{figure}%\vspace{-0.2cm}

In this formulation, which follows and unifies prior work on incentivized exploration, in each round $t\in [T]$ a new agent $t$ arrives with some type $\type_t\in\types$ and can choose an action from a finite set $\arms$ of actions (arms).\footnote{We identify it as $\arms = [K]$, where $K = |\arms|$ is the number of arms, unless specified otherwise.} Upon arrival, a \emph{public type} $\pubtype_t = \pub(\type_t)$ that corresponds to some mapping $\pub:\types\to\pubtypes$ is immediately observed by the principal. The principal then generates a message $\msg_t$ according to a pre-determined policy $\pi$ that is measurable with respect to the history of all actions and observations thus far. Having observed this message, the agent selects an arm $\arm_t\in\arms$ and subsequently collects a reward $\rew_t$. The principal observes the chosen arm $\arm_t$, the reward $\rew_t$, and possibly also some problem-specific auxiliary information $\aux_t\in \AUX$.

\begin{remark}
(Information Asymmetry.) Note that while the principal has access to the full history of observations, each agent $t$ observes only what the principal chooses to reveal in its message $\msg_t$. On the other hand, the principal can only send messages, whereas the reward-generating actions are controlled by the agents.
\end{remark}

The reward and auxiliary feedback are drawn independently from some distribution $\outD$ parameterized by the model, the agent type, and the chosen arm: $(\rew_t,\aux_t)\sim \outD(\trueModel,\type_t,\arm_t)$.
Thus, $\outD$ is a fixed and publicly known mapping from $\MODELS\times\types\times\arms$ to distributions over $\R\times\AUX$, to which we refer as the \emph{outcome distribution}. The corresponding marginal distribution for rewards is denoted by $\rewD$, that is, $\rew_t\sim \rewD(\trueModel,\type_t,\arm_t)$.
For convenience, the expected reward is denoted by $\rewE(\model,x,i) := \E\sbr{\rewD(\model,x,i)}$,
for any fixed $(\model,x,i)\in \MODELS\times\types\times\arms$.\footnote{In fact, our analysis assumes a more general outcome specification whereby only $\E\sbr{ (\rew_t, \aux_t) \mid \trueModel = \model,\type_t = x,\arm_t = i}$, for any fixed $(\model,x,i,t)\in \MODELS\times\types\times\arms\times\N$, is known, and the corrupting process is ``well-behaved'' (see Assumption~\ref{assumption:noise}).}

\xhdr{Bayesian setting.} Before the game starts, "nature" draws (an unknown) model $\trueModel\in\MODELS$ according to a publicly known Bayesian prior $\prior$. Further, the principal announces its messaging policy $\pi$ and message set $\msgs$, \ie commits to how she will generate messages given her observations. Crucially, the agents observe this commitment, and the principal cannot deviate from it during the game.\footnote{We emphasize the distinction between \emph{problem structure}, which is known to everyone, from \emph{problem instance}, which is not known to anyone. In our problem, the problem structure consists of the primitives  from Figure~\ref{fig:IE-protocol}, namely the sets
    $\arms,\MODELS,\AUX,\types,\pubtypes$, horizon $T$, and the mapping $\pub$,
as well as outcome distribution $\outD$ and prior $\prior$. Within this structure, a problem instance further specifies the model $\trueModel\in\MODELS$ and the type sequence.}

\xhdr{Objectives.} Each agent aims to maximize its expected reward (see Assumption~\ref{assn:rationality} for a formal specification). The principal strives to maximize the total reward $\sum_{t\in[T]} \rew_t$.

\xhdr{Types and menus.} In terms of agent types, there are three canonical scenarios: \emph{homogeneous agents} that are all of the same type (\ie $|\types|=1$), \emph{public types} that are completely observable (\ie $\pub()$ is the identity mapping), and \emph{private types} that are unobserved (formally: $|\pubtypes|=1$).

We stipulate that each principal's message $\msg\in\msgs$ implies a "menu" of direct recommendations. Formally, a mapping $\menu:\types\times\msgs\to\arms$, called \emph{recommendation menu}, is announced as a part of the messaging policy $\pi$, such that each agent $t$ can infer the direct recommendation as $\menu(\type_t,\msg_t)$.

\begin{remark} (Special problem formulations) If each agent $t$ follows recommendations, $\arm_t = \menu(\type_t,\msg_t)$, our problem boils down to a contextual bandit problem (with public types as contexts). The case of homogeneous agents and no auxiliary feedback corresponds to the stochastic bandit problem, the classical bandit setting with IID rewards. %Exploration is needed since the after-round feedback does not include counterfactual rewards for all arms.
\end{remark}

\xhdr{Bayesian Incentive-Compatibility.} Let us spell out the requisite \emph{Bayesian Incentive-Compatibility} property, henceforth abbreviated as \emph{BIC}. Following most prior work on incentivized exploration, we endow the principal with a \emph{warm-start}: the first $T_0$ rounds (for some fixed and known $T_0$) that take place exogenously and are not subject to incentives. The data collected during the warm-start is referred to as \emph{warm-up data} and denoted by $\warmup = \rbr{ \rbr{\type_t,\arm_t,\rew_t,\aux_t}:\; t\in[T_0] }$. This data is collected by some mechanism (whether within the framework of incentivized exploration or not), and is used by the principal to create incentives afterwards. Specifically, we will guarantee BIC for all rounds $t>T_0$ under some assumption on $\warmup$; we refer to the collection of rounds $t>T_0$ as the \emph{main stage}.

The BIC property states that each agent in the main stage prefers to follow the recommended arm.\footnote{Agent's preferences are in terms of expected rewards, which are well-defined only conditional on a particular type sequence, the realization of $\warmup$, and the model of agents' behavior for all previous agents. By a slight abuse of notation, the expectation $\E[y_t \mid \cal{E}]$ is well-defined as long as conditioning on event $\cal{E}$ yields a valid probability space; this holds even when $\cal{E}$ is itself \emph{not} included in our probability space.} Formally, the BIC property is defined as follows.

\begin{definition}\label{def:BIC}
(BIC) Let $\cal{E}_t$ denote the event that all agents before round $t$ followed recommendations:
$\arm_s = \menu(\msg_s,\type_s)$ for all rounds $s$ with $T_0<s<t$. The messaging policy is \emph{BIC} for a given realization of the warm-up data,
    $w\in \rbr{\types\times\arms\times\R\times\AUX}^{T_0}$,
if the following holds:
\begin{align}
&\E\sbr{\rew_t \mid \cal{E}_t,\;
    \vec{\type}=\vec{x},\;
    \warmup = w,\;
    \msg_t =m,\;
    \arm_t = \menu(x_t,m) } \nonumber \\
&\qquad\qquad-
\E\sbr{\rew_t \mid \cal{E}_t,\;
    \vec{\type}=\vec{x},\;
    \warmup = w,\;
    \msg_t =m,\;
    \arm_t = i }
\geq 0,
    \label{eq:BIC-defn}
\end{align}
for all rounds $t>T_0$, all type sequences $\vec{x} = (x_1 \LDOTS x_T)\in \types^T$ consistent with $w$, all messages $m\in\msgs$ such that
$\Pr\sbr{\msg_t=m \mid \cal{E}_t,\; \vec{\type}=\vec{x},\;\warmup = w}>0$,
and all arms $i\in\arms$. The expectation in \refeq{eq:BIC-defn} is taken over all applicable randomness: in the choice of the model, in the messaging policy, and in the outcome distribution.
\end{definition}

\begin{remark} (Non-Compliance) Conditioning BIC on the compliance event $\cal{E}_t$ is standard in prior work on incentivized exploration. However, all our results carry over to a setting in which $\cal{E}_t$ is replaced by any other model of agent behavior, compliant or not. This is because our BIC guarantees rely on $\warmup$ alone, regardless of the data collected in the main stage. The analyses carry over word-by-word, and we condition on $\cal{E}_t$ for ease of presentation only.
\end{remark}

We also consider a more refined version of the BIC property. For a fixed $\eps\in \R$, the messaging policy is called \emph{$\eps$-BIC} if \refeq{eq:BIC-defn} holds with $\eps$ on the right-hand side. We call this \emph{$\eps$-strong-BIC} if $\eps>0$, and $|\eps|$-weak-BIC if $\eps<0$. In some scenarios one may ensure $\eps$-strong-BIC, providing for stronger incentives. In some other scenarios, one may only ensure $\eps$-weak-BIC for any given $\eps>0$ (with an assumption on the warm-up data that depends on $\eps$). Unless mentioned otherwise, we assume a relaxed version of rationality for agents' behavior:

\begin{assumption}\label{assn:rationality}
Agents follow recommendations for any $\eps_0$-BIC policy, for some fixed $\eps_0<0$.
\end{assumption}

Then, it suffices to ensure $\eps$-BIC for any $\eps\geq \eps_0$. We emphasize that all results with $\eps$-BIC, $\eps\geq 0$, accommodate the "standard" version of agent rationality with $\eps_0=0$.

\xhdr{The learning problem: linear contextual bandits.} We consider the \emph{linear contextual bandit} setting \citep[\eg][]{Langford-www10,Reyzin-aistats11-linear,Csaba-nips11}
as the general learning problem. We next define this problem in terms of our existing structure. The model set is $\MODELS\subset \R^d$ for some fixed dimension $d\geq 1$. The types are captured by $K\times d$-dimensional matrices, where $K$ is the number of arms: $\types\subset \R^{K \times d}$. We write each type $x\in \types$ as $x = (x_1 \LDOTS x_K)$, where $x_i\in \R^d$ is the feature vector for each arm $i\in\arms$. Likewise, we write the realized types as $\type_t = \rbr{\type_{1,t} \LDOTS \type_{K,t}}$, where
$\type_{i,t}\in \R^d$ for each arm $i\in\arms$. The expected rewards are linear:
\begin{align}
    \rewE(u,x,i) = x_i\cdot u, \quad \text{for all models $\model$, types $x$ and arms $i$}.
\end{align}

In each round $t$, the realized rewards are generated as follows. First, the round-$t$ noisy model $r_t  = \trueModel + \xi_t\in \R^d$ is realized, where $\xi_t\in \R^d$ is a random \emph{noise vector} with zero mean. Then the reward of a given arm $i$ is generated as $\type_{i,t}\cdot r_t$. The noise vectors are independent across coordinates and time periods, and in each coordinate the noise is mean-$0$ sub-Gaussian with variance at most $R^2$. In fact, our analysis allows for a more general noise shape, whereby the sub-Gaussian noise can be correlated across dimensions and time periods.\footnote{A sufficient assumption is as follows. Let $\cal{F}_t$ be the sigma-algebra generated by all of the principal's observations during all rounds up to and including round $t$.
%    $\cal{F}_{t-1}:= \sigma\left( {\type_s}, \arm_s,Y_s: s\in[t-1]\right)$.
The zero-mean noise process $\rbr{ {\xi}_t \in\R^d : t = 1,2,... }$ must satisfy \[ \left. w\cdot {\xi}_t \;\right\rvert\;
    \sigma\rbr{ {\type_t}, \arm_t, \cal{F}_{t-1}, \trueModel} \sim \subG\rbr{\norm{w}_2^2\; R^2}
\quad\text{for each round $t\in\N$ and any $w\in\sigma\rbr{ {\type_t}, \arm_t, \cal{F}_{t-1}, \trueModel}$.}
\]\vspace*{-0.2in}} 
% see Assumption~\ref{assumption:noise} for a more general noise shape.

We consider two types of partial feedback that are common in the bandits literature: \emph{bandit feedback} ($\aux_t = \emptyset$) and \emph{semi-bandit feedback}, where
    $\aux_t \subseteq \cbr{r_t\cdot\e_j : j\in[d]}$ with $\e_j$ denoting the $\ith{j}$ standard basis vector of $\R^d$.\footnote{Defining realized rewards via $r_t$ is somewhat non-standard; we do it to include both bandit and semi-bandit feedback in a unified way. A more standard definition adds scalar noise directly to expected reward; our analysis carries over.} The case of homogeneous agents is "embedded" by fixing dimension $d=K$ and type set $\types = \cbr{\cal{I}_K}$.

\xhdr{Spectral diversity.} We quantify the diversity of the principal's data via the smallest eigenvalue of the empirical covariate matrix.%
\footnote{This is common in linear contextual bandits \cite[\eg][]{Csaba-nips11,goldenshluger2011note}.}
For a subset $\cal{T}\subset [T]$ of rounds, define
\begin{align}
\emin[\cal{T}] := \lambda_{\min}\rbr{\hat{\Sigma}_{\cal{T}}},
\text{ where }
\hat{\Sigma}_{\cal{T}} = \textstyle \sum_{t\in\cal{T}} \type_{\arm_t,t}\otimes \type_{\arm_t,t}
\end{align}
is the empirical covariate matrix and $\lambda_{\min}(\cdot)$ is the smallest eigenvalue. We call $\emin[\cal{T}]$ the \emph{spectral diversity} of the data from $\cal{T}$. We are particularly interested in $\emin$, the spectral diversity of the warm-up data. The sufficient assumption for all our results, in terms of the warm-up data, is that $\emin$ is sufficiently large. It is known that the spectral diversity can only increase with more data:
   $\emin[\cal{T}]\leq \emin[\cal{T}^{\,\prime}]$ whenever $\cal{T}\subset \cal{T}^{\prime} \subset [T]$.

It is known that near-uniform exploration (\ie sampling each arm with at probability at least $\eps/K$, for some $\eps>0$) achieves spectral diversity $\emin$ that is (at least) linear in $\nicefrac{\eps}{K}\cdot T_0$ (This holds under some assumptions on the context arrivals, see Lemma~\ref{lemma:near-uniform-exploration}.) Thus, our lower-bound assumptions on $\emin$ can be satisfied by near-uniform exploration in sufficiently many rounds of the warm-start stage, so this is one reasonable intuition for these assumptions.

We also consider a closely related notion, $\eminprime[\cal{T}] := \lambda_{\min} \rbr{\hat{\Sigma}_{\cal{T}} \odot\cal{I}_d}$, where "$\odot$" is the Hadamard product and $\cal{I}_d$ denotes an identity matrix in $\R^{d\times d}$. It is known that
    $\eminprime[\cal{T}] \geq \emin[\cal{T}]$ whenever $\emin[\cal{T}] > 0$ \cite{ando1995majorization}.

\xhdr{Other preliminaries.} We make some (fairly common) assumptions on the parameter sets $\MODELS,\types$.
\begin{assumption}[model set $\MODELS$ and type set $\types$]
\label{assumption:parameter-set}
For some absolute constants $C_{\MODELS},C_{\types},\s>0$,
~%The model space $\cal{U}$ and the type space $\cal{X}$ satisfy the following properties:
\begin{enumerate}
\item $\sup_{\model\in\MODELS}\norm{\model}_2 \leq C_{\MODELS}$.
\item Each row $i\in\cal{K}$ of every type $x\in\cal{X}$ satisfies
    $\norm{x_i}_0 \leq \s$
and
    $\norm{x_i}_2 \leq C_{\types}$.
\end{enumerate}
\end{assumption}

We say the warm-start data is "non-adaptive" if the action sequence $\rbr{\arm_t : t \in[T_0]}$ is selected before round $1$. In particular, it is independent of the noise sequence $\rbr{\xi_t : t\in[T_0]}$. Some of our guarantees strengthen when this is the case. To simplify the notation, we let $\E_0[\cdot]$ and $\Prob[0]{\cdot}$ denote, respectively, expectation and probability over the random draw from the prior, $\trueModel\sim\prior$. Symbols "$\gtrsim$" and "$\lesssim$" denote "$\geq$" and "$\leq$" respectively, up to multiplicative factors that are positive universal constants.

%% file: main-result-section.tex
We next present our algorithm, called \emph{\ourAlg}, and formulate the main results: the general guarantee and the main corollaries pertaining to private types, informative recommendations, and correlated priors. The general guarantee comprises the common part of all our analyses (with some last steps being scenario-specific), and the corollaries represent the main "dimensions" along which we extend prior work. Additional corollaries are relegated to Section~\ref{sec:TS-other}.

%We make some (fairly common) assumptions on the parameter sets $\MODELS,\types$.
%\begin{assumption}[model set $\MODELS$ and type set $\types$]
%\label{assumption:parameter-set}
%For some absolute constants $C_{\MODELS},C_{\types},\s>0$,
%~%The model space $\cal{U}$ and the type space $\cal{X}$ satisfy the following properties:
%\begin{enumerate}
%\item $\sup_{\model\in\MODELS}\norm{\model}_2 \leq C_{\MODELS}$.
%\item Each row $i\in\cal{K}$ of every type $x\in\cal{X}$ satisfies
%    $\norm{x_i}_0 \leq \s$
%and
%    $\norm{x_i}_2 \leq C_{\types}$.
%\end{enumerate}
%\end{assumption}

%\ascomment{TODO: explain}

%\begin{assumption}[$R$-sub-Gaussian Noise Process]
%\label{assumption:noise}
%Define $\cal{F}_{t-1}:= \sigma\left( {\type_s}, \arm_s,Y_s: s\in[t-1]\right)$. The zero-mean $\R^d$-valued process $\left( {\xi}_t : t = 1,2,... \right)$ satisfies the following for each $t\in\N$:
%For any fixed $w\in\R^d$,
%%For any $w\in\R^d$ that is conditionally independent of ${\xi}_t$ given
%%$\rbr{{\type_t}, \arm_t, \cal{F}_{t-1}, \trueModel}$, one has that
%\[ \left. w^\top{\xi}_t \;\right\rvert\;
%    \sigma\rbr{ {\type_t}, \arm_t, \cal{F}_{t-1}, \trueModel,w} \sim \subG\rbr{\norm{w}_2^2\; R^2}.\]
%\end{assumption}

%We start with the general Theorem~\ref{thm:general} essentially outlines the general conditions under which one could hope to achieve BIC using \ourAlg.

\xhdr{Semantics-consistent policies.} We focus on messaging policies of a certain form that endows messages with a consistent semantics. In each round $t$, the policy selects some model $\model_t\in\MODELS$
(as an educated guess which trades off exploration and exploitation), and generates a message determined by this model and the public type $\pubtype_t$. Specifically,
    $\msg_t = \smap(\pubtype_t, \model_t)$,
where $\smap:\pubtypes\times\MODELS\to\msgs$ is some fixed and known mapping called \emph{semantic map}. Thus, the pair $(\msgs,\smap)$ specifies message semantics relative to the chosen model. This pair be fixed exogenously or chosen endogenously, depending on a particular scenario.
%We call such messaging policies \emph{semantics-consistent}, summarizing them
We summarize this messaging policy in Figure~\ref{fig:semantics}.%\vspace{-0.3cm}

\begin{figure}[ht]
\centering
\caption{``Semantics-consistent" messaging policy}
\begin{tabular}{|p{\textwidth}|}
\hline\vspace{0.1mm}
\emph{Announce:}
message set $\msgs$,
semantic map $\smap:\pubtypes\times\MODELS\to \msgs$,
and model-selection policy $\pimod$. \\
In each round $t = T_0+1,\, T_0+2 \LDOTS T$:
\begin{enumerate}
  \item use past observations to generate model $\model_t\in\MODELS$, according to policy $\pimod$.
  \item generate message for agent $t$ as $\msg_t = \smap(\pubtype_t,\model_t)$.
\end{enumerate}
\\\hline
\end{tabular}
\label{fig:semantics}
\end{figure}
\vspace*{-0.2in}
We require the semantic map $\smap$ to provide a consistent recommendation menu:

\begin{definition}[Menu-consistency]\label{assn:consistency}
Consider semantic map $\smap$ such that there exists some
    $\menu_{\smap}: \types\times\msgs\to\arms$
such that the following holds: for each type $x\in\types$, each message $m\in\msgs$, and each model $\model\in\MODELS$ that maps to this message, $m=\smap(\pub(x),\model)$, \begin{align}\label{eq:consistent-defn}
\rewE(\model,x,i) -\rewE(\model,x,j) \geq 0
\quad \text{for arm $i=\menu_{\smap}(x,m)$ and each arm $j\in\arms$}.
\end{align}
We call such $\smap$ \emph{menu-consistent}, and associate it with one such menu $\menu_{\smap}$. We call $\smap$ \emph{$\alpha$-menu-consistent} if \refeq{eq:consistent-defn} holds with $\alpha \in \R$ on the right hand-side. To indicate $\alpha \geq 0$, we use \emph{$\alpha$-strong menu-consistent} for $\smap$; likewise for $\alpha<0$, we use \emph{$\mod{\alpha}$-weak menu-consistent}.
\end{definition}

\begin{remark}\label{rem:consistency}
If $\smap$ is menu-consistent, then $\menu_{\smap}$ is consistent (up to tie-breaking) with the optimal (model-aware) menu $\menu^*(x,\model) = \textstyle \argmax_{i\in\arms} \rewE(\model,x,i)$, which returns the best arm for each type-model pair
    $(x,\model)\in \types\times\MODELS$.
Specifically, fixing $(x,\model)$ and message $m = \smap(\pub(x),\model)$, arm
    $i = \menu_{\smap}(x,m)$
maximizes expected reward
    $\rewE(\model,x,\cdot)$. If $\smap$ is $\alpha$-menu-consistent with some $\alpha>0$, then $\menu_{\smap}$ is unique and coincides with $\menu^*$.

For public types (incl. the special case of homogeneous agents), the standard choice of the semantic map $\smap$ is the optimal menu: $\smap = \menu^* = \menu_{\smap}$. However, other choices of $\smap$ may be useful to handle private types and/or provide more informative recommendations.
\end{remark}

% We sometimes consider another version of menu-consistency, as a sufficient condition.

% \begin{definition}\label{defn:alpha-consistency}
% Semantic map $\smap$ is called \emph{$\alpha$-consistent}, $\alpha\geq 0$, if the following holds: for each type $x\in\types$ and each message $m\in\msgs$, there is \emph{some} model $\model\in\MODELS$ that maps to this message, $m=\smap(\pub(x),\model)$, such that
% \refeq{eq:consistent-defn} holds with $\alpha$ on the right-hand side.
% \end{definition}

\xhdr{Our Algorithm: \OurAlg.}
We consider a semantic-consistent messaging policy where the model $\model_t$ in each round $t$ is generated by posterior sampling. That is, the policy computes Bayesian posterior $\cal{P}_t$ given the current observations, and draws $\model_t$ independently from this posterior. The message is then "filtered" via the semantic map $\smap$. We call this algorithm \emph{\ourAlg}. When $\smap = \menu^*$, this algorithm coincides with Thompson Sampling, a standard approach in multi-armed bandits. For the general $\smap$, $\menu_{\smap}$ is consistent with recommendation menu induced by Thompson Sampling as long as $\smap$ is menu-consistent, as per Remark~\ref{rem:consistency}.

\begin{remark}
Menu-consistency of $\smap$ is needed to ensure consistency with Thompson Sampling, not to prove incentive-compatibility.
\end{remark}

\xhdr{Primitives.}
We express our guarantees in terms of the following primitives:
for each type $x\in\types$ and message $m\in\msgs$ and arms $i,j\in\arms$,
\begin{subequations}
\label{eqn:primitives}
\begin{align}
%&\menuset_{\smap}(x,m) :=
%    \bigcup_{\text{models }\model \in \MODELS:\;\; \smap(x,\model) = m}\quad
%    \argmax_{\text{arms } i\in\cal{K}}\; \rewE(u,x,i)
%    \label{eqn:primitives4}\\
&\Delta_{i,j}\pbrac{x,m} := \E_0
    \sbr{\rewE(\trueModel,x,i) - \rewE(\trueModel,x,j) \mid \smap\rbr{\pub(x),\trueModel} = m },
        \label{eqn:primitives2} \\
&\delTS := \inf_{x\in\cal{X}}\min_{m\in\cal{M}}\Prob[0]{\smap\rbr{\pub(x),\trueModel} = m}. \label{eqn:primitives3}
\end{align}
\end{subequations}
Thus, $\Delta_{i,j}\pbrac{x,m}$  is the expected "gap" between arms $i$ and $j$ for a given type-message pair, and $\delTS$ is the smallest message probability over all type-message pairs.

\begin{theorem}[General Guarantee]
\label{thm:general}
%Consider a menu-consistent semantic map $\smap$ and
Assume that $\delTS>0$, as per \refeq{eqn:primitives3}.
%Suppose Assumptions \ref{assn:consistency}, \ref{assumption:parameter-set}, \ref{assumption:noise} hold, and that $\delTS$ (as defined in \refeq{eqn:primitives3}) is positive.
Fix $\eps>0$ and suppose that the spectral diversity of the warm-up data satisfies
    $\emin \gtrsim \Lambda(\eps) := \pbrac{D/\eps^2}\log\pbrac{2/\delTS}$
almost surely, where $D$ is specified below. Further, suppose one of the following holds:
\begin{enumerate}

    \item Warm-up data is non-adaptive, has bandit feedback, and $D = C_{\cal{X}}^2R^2$.

    \item Warm-up data has bandit feedback, and $D = d\pbrac{RC_{\cal{X}} + C_{\cal{U}}}^2\log\pbrac{C_{\cal{X}}^2T+3}$.

    \item Warm-up data has semi-bandit feedback, type set satisfies $\cal{X} \subseteq \cbrac{0,1}^{K \times d}$, and $D = \s R^2$.
\end{enumerate}

Then, \ourAlg
%in any round $t\in\cbrac{T_0+1, \dots , T}$
is $g(\eps)$-BIC, with %$g(\eps)$ given by 
\begin{align}
g(\eps) :=
    \inf_{x\in\cal{X}}\min_{m\in\cal{M}}
    \min_{j\in\cal{K}\backslash \cbr{i} }\pbrac{\Delta_{i,j}\pbrac{x,m} - \tfrac{\eps}{4}\,\norm{x_{i} - x_{j}}_2},
    \quad \text{where } i= \menu_{\smap}(x,m).
     \label{eqn:slack}
\end{align}

\end{theorem}

\begin{remark}
    For part (3), we only need a weaker condition:
    $\eminprime \gtrsim \Lambda(\eps)$.
     %to hold, ceteris paribus. This is milder since $\eminprime \geq \emin$ for positive-definite matrices \cite{ando1995majorization}.
\end{remark}

%Proof is provided in \S\ref{proof:main}. 
The takeaway is that as $\emin$
%the spectral diversity of the warm-up data
increases, it becomes progressively more incentive-compatible for agents to follow principal's recommendations. More precisely: as $\emin$ increases, one can choose smaller $\eps$ so that $\emin\geq \Lambda(\eps)$ still holds, and then $g(\eps)$ decreases.
%This is reflected via monotone increments in $g(\eps)$ as $\eps$ vanishes.
%Of note here is that the initial data could be collected using \text{any} algorithm, or simply purchased for a monetary payment. The mode of initial data collection will not affect the mechanism's round $t$ guarantees provided in Theorem~\ref{thm:general} as long as a relevant eigenvalue condition is met (except in scenario 1 which posits that the data is collected, essentially, \textit{non-adaptively}).

Going forward, we specialize Theorem~\ref{thm:general} for particular scenarios. For each scenario, we analyze $g(\eps)$ and choose $\eps$ and (possibly) the semantic map $\smap$ to ensure that $g(\eps)$ is sufficiently low, namely $g(\eps)\leq \eps_0$ in line with  Assumption~\ref{assn:rationality}.

%relevant message spaces in several novel as well as canonical test cases for incentivized exploration: (i) linear bandits with private types; (ii) stochastic bandits with informative recommendations; (iii) sleeping bandits with unobservable feasible arm-subsets; (iv) combinatorial semi-bandits; and (v) standard linear contextual bandits. In all of the aforementioned paradigms, our approach will be to set design parameters (choice of $\eps$ and the semantic mapping $\smap:\cal{X}\times\cal{U} \mapsto \cal{M}$) such that that $g(\eps)$ in absolute value remains below the target $\mod{\eps_0}$.

%% file: implications.tex
%\subsection{Linear bandits with private agent types}
\subsection{Private agent types}
\label{subsection:CBprivate}

We next turn our attention to analyze private agent types. Note that there is only one public type, call it $\xpub$, and one may write $\smap(\xpub,\model) = \smap(\model)$.

\begin{corollary}[private types]
%[\OurAlg is $\eps_0$-BIC for bandits with private types]
\label{corollary:CB-private}
%Suppose Assumptions~\ref{assumption:parameter-set} and \ref{assumption:noise} hold.
Fix $\eps>0$ and consider the setting in Theorem~\ref{thm:general}, specialized to private types.  Suppose semantic map $\smap$
% $\smap: \pubtypes \times \cal{U}\mapsto \cal{M}$
%is $\alpha$-consistent for some $\alpha > \mod{\eps_0}/2$
%(see Definition~\ref{defn:alpha-consistency}),
is $\alpha$-menu-consistent for some $\alpha \in \R$. Then \ourAlg is $\rbr{\alpha-\eps/2}$-BIC.
%is $\alpha$-consistent for $\alpha=\eps/2$ and $\rho$-granular for some
%    $\rho \leq \eps/\rbr{4\,C_{\types}}$.
%Then, Assumption~\ref{assn:consistency} is satisfied,
%Then $\smap$ is menu-consistent, and \ourAlg is $\rbr{\alpha-\eps}$-BIC.

%If $\delTS$ (as defined in \eqref{eqn:primitives-private-delta}) is positive, spectral diversity of the warm-up data satisfies $\emin \gtrsim \pbrac{DC_{\types}^2/\eps_0^2}\log\pbrac{2/\delTS}$ w.p. $1$, and one of the three conditions given in Theorem~\ref{thm:general} is met, then \ourAlg in any round $t\in\cbrac{T_0+1, \dots, T}$ is $\pbrac{\alpha+\eps_0}$-BIC.

%Further, if $\smap$ is $\alpha$-consistent for some $\alpha > \mod{\eps_0}/2$ (see Definition~\ref{defn:alpha-consistency}), then there exists a unique plausible best arm for each type and each message, i.e., $\mod{I_{{x}}(m)} = 1\ \forall\ x\in\types,\ m\in\cal{M}$, and \ourAlg in round $t$ is $\pbrac{\alpha+\eps_0}$-BIC.

\end{corollary}

\begin{remark}
If $C_{\types}$ from Assumption~\ref{assumption:parameter-set} is \emph{not} treated as an absolute constant, then Corollary~\ref{corollary:CB-private} requires a scale-invariant (and slightly larger) lower bound on spectral diversity:
    $\emin \gtrsim C_{\types}^2\cdot \Lambda(\eps)$.
\end{remark}
%\akcomment{$\alpha$-consistency and $\rho$-granularity for this choice of parameters ensures menu-consistency \`a la Assumption~\ref{assn:consistency}, which allows us to invoke Theorem~\ref{thm:general}}.

\begin{remark}
While we put the spotlight on private types, we note that Corollary~\ref{corollary:CB-private} applies to various semantic maps $\smap$ and allows correlated priors. 
\end{remark}

%Proof is provided in \S\ref{proof:corollary-private}.

%\ascomment{TODO: examples!!}

%\akcomment{See below..}

We next provide some examples with specific message sets, semantic maps and associated menus, together with resulting guarantees. A common feature of these examples is that the nature of menu-consistency and BIC guarantees (weak or strong) is intrinsically tied to the geometric properties of the model and type sets when types are private. In contrast, with public types (i.e., when $\pubtypes=\types$), one can always guarantee menu-consistency irrespective of the problem geometry. With private types, the model and type set geometries must be conducive for a menu-consistent semantic map to exist and for BIC to be feasible, as forthcoming examples illustrate.

%To derive a clean BIC property, we need $\smap$ to be sufficiently ``granular":

\begin{enumerate}

    \item \textbf{Weak menu-consistency and weak BIC.} Fix $\eps>0$ and an $\eps/\pbrac{4C_{\cal{X}}}$-cover w.r.t. $\ell_2$ norm (see Definition~\ref{definition:eps-cover}) of the model set $\cal{U}$ as the message set $\cal{M}$; call it $\scr{U}$, i.e., $\cal{M} = \scr{U}$. Denote by $\cbr{\scr{P}_{\model} : \model \in \scr{U}}$ the Voronoi partition (see Definition~\ref{definition:voronoi}) of $\MODELS$ generated using elements of $\scr{U}$ as centers (or seeds). Then, note that each $\scr{P}_{\model}$ (for $\model\in\scr{U}$) is $\eps/\rbr{4C_{\cal{X}}}$-granular (see Definition~\ref{definition:granular}). Define the semantic map $\smap: \MODELS \mapsto \scr{U}$ by
    \begin{align*}
        \smap(\model) = \sum_{\model^\prime\in\scr{U}}\model^\prime\indi{\model\in\scr{P}_{\model^\prime}}.
    \end{align*}
    Associate with this semantic map the menu $\menu_\smap: \types \times \scr{U} \mapsto \cal{K}$ given by
    \begin{align*}
        \menu_\smap(x,\model) = \argmax_{i\in\cal{K}}x_i \cdot  \model,
    \end{align*}
    with ties broken according to some fixed rule. Then, $\smap$ is $\pbrac{-\eps/2}$-menu-consistent, and consequently, \ourAlg is $\rbr{-\eps}$-BIC (by Corollary~\ref{corollary:CB-private}). Note, however, that $\menu_\smap$ may not be consistent with the optimal model-aware menu $\menu^*$ because of weak menu-consistency of $\smap$. To achieve a stronger notion of menu-consistency (and thereby, consistency with $\menu^*$), geometric assumptions on the model and type sets are necessary, as forthcoming examples illustrate.

    \item \textbf{Menu-consistency and weak BIC.} In the previous example, if the model set were simply given by the collection of Voronoi centers (i.e., $\cal{U} = \scr{U}$), then $\smap$ would be menu-consistent and as a result, \ourAlg $\pbrac{-\eps/2}$-BIC. In this case, finiteness of the model set $\MODELS$ allows lifting the weak menu-consistency guarantee up to its stronger counterpart.

    \item \textbf{Strong menu-consistency and strong BIC.} The sleeping bandits paradigm where the message set is the set of all rankings over arms, and the semantic map induces a menu of ``top feasible arms,'' satisfies strong menu-consistency as well as BIC properties under mild assumptions on the prior (a detailed discussion is deferred to \S\ref{subsubsection:sleeping-bandits}).

    \item \textbf{Strong menu-consistency and strong BIC.} Fix a collection of $N \geq K$ equi-separated points lying in the interval $[-1,1]$; call it $V = \cbrac{v_i : i\in[N]}$. The type set $\cal{X}$ is given by the collection of vectors in $\R^K$ whose coordinates are distinct elements of $V$.  The model set is initialized to $\cal{U} = \sbrac{-1,-\delta}\cup\sbrac{\delta,1}$ for some $\delta\in\pbrac{0, 1}$. Observe that $\min_{\model\in\MODELS}\min_{i,j\in[N]: i\neq j}\mod{\pbrac{v_i - v_j}\cdot\model} = \Omega\pbrac{\delta/N}$. Set $\cal{M} = \cbrac{-1,+1}$ as the message space, and define the semantic map $\smap: \MODELS \mapsto \cbrac{-1,+1}$ given by $\smap(\model) = \model/\mod{\model}$. Associate with this $\smap$ the menu $\menu_\smap: \types \times \cbrac{-1,+1} \mapsto \cal{K}$ given by $\menu_\smap(x,m) = \argmax_{i\in\cal{K}}x_i \cdot m$, with ties broken according to some fixed rule. Such $\smap$ is $\Omega\pbrac{\delta/N}$-menu-consistent. As a result, \ourAlg needs $\varepsilon$ of order $\delta/N$, or equivalently, spectral diversity $\emin$ of order $N^2/\delta^2$ to achieve strong BIC in this setting.

    \item \textbf{Strong menu-consistency and strong BIC.} Fix $N \geq 2$ and consider $2N \geq K$ equi-separated points on the surface of the unit ball in $\R^2$; call this set $V = \cbrac{v_i : i\in[2N]}$. The type set $\cal{X}$ is given by the collection of $\pbrac{K\times 2}$-dimensional matrices whose rows are distinct elements of $V$. The model set is initialized to $\MODELS = \cbrac{\model \in \R^2 : \norm{\model}_2 = 1, \quad \min_{i,j\in[2N]: i\neq j}\mod{\pbrac{v_i - v_j}\cdot \model} \geq 1/\pbrac{CN}^2}$, where $C$ is some large enough universal constant (specifically, $C$ is large enough to ensure that $\MODELS$ is non-empty and can be represented as a disjoint union of $4N$ connected components). Then, any semantic map $\smap: \MODELS \to \cal{M}$ that maps models in the same connected component of $\MODELS$ to the same message in $\cal{M}$, is $\Omega\pbrac{1/N^2}$-menu-consistent. Thus, \ourAlg needs $\varepsilon$ of order $1/N^2$, or equivalently, spectral diversity $\emin$ of order $N^4$ to achieve strong BIC in this setting.

\end{enumerate}

%\textbf{Examples.}  \akcomment{See the examples following Corollary~\ref{corollary:bandits-informative-recos}}.

% \begin{remark}
% If $\cal{U} = \cbrac{u \in \R^d : \norm{u}_2 \leq 1}$, each $U(m)$ contains some $\ell_2$ ball of radius $\varepsilon/C$ (for some universal constant $C>0$),  and $\cal{P}_0$ is Uniform over $\cal{U}$, then $\delTS \geq \pbrac{\varepsilon/C}^d$.  In this case,  $\lambda_{\min}\pbrac{\hat{\Sigma}_{\cal{T}}} \gtrsim \frac{dC_{\types}^4R^2}{\varepsilon^2}\log\pbrac{\frac{C}{\varepsilon}}$ suffices to achieve $\varepsilon$-BIC.  Naturally, the claim also extends to the standard setting with public types, i.e., when the sequence $\pbrac{{{x}^{(t)}}: t=1,2,...}$ is observable.

% \end{remark}

%\subsection{Stochastic K-armed bandits with informative recommendations}
\subsection{Informative recommendations}
\label{subsection:informative-recommendations}

This subsection zooms in on informative recommendations, as expressed by  exogenously given semantic map $\smap$. To showcase our results along this ``problem dimension", let us keep the rest of the problem simple, focusing on homogeneous agents (\ie with unique type $\xsingleton$).

%We restate below the primitives (from \eqref{eqn:primitives}) for the simplified semantic map for this setting.
%
%\begin{subequations}
%\label{eqn:primitives-bandits-informative-recos}
%\begin{align}
%&\menuset_{\smap}(\xsingleton,m) :=
%    \bigcup_{\text{models }\model \in \MODELS:\;\; \smap(\xsingleton, \model) = m}\quad
%    \argmax_{\text{arms } i\in\cal{K}}\; \model_i,  \label{eqn:primitives-bandits-informative-recos-plausible-best} \\
%&\Delta_{i,j}\pbrac{\xsingleton,m} := \E_0
%    \sbr{\trueModel_i - \trueModel_j \mid \smap\pbrac{\xsingleton, \trueModel} = m }, \label{eqn:primitives-bandits-informative-recos-Delta} \\
%&\delTS := \min_{m\in\cal{M}}\Prob[0]{\smap(\xsingleton, \trueModel) = m}.  \label{eqn:primitives-bandits-informative-recos-delta}
%\end{align}
%\end{subequations}
%Here,  message $m\in\cal{M}$ and arms $i,j\in\cal{K}$.

% \begin{assumption}[Informative recommendations]
% \label{assumption:informative-recommendations}
%     For each message $m\in\cal{M}$, there exists a unique plausible best arm in $I(m)$, i.e., $\mod{I(m)} = 1\ \forall\ m\in\cal{M}$.
% \end{assumption}

\begin{corollary}[exogenous $\smap$ and homogeneous agents]
%[\OurAlg is BIC for bandits with informative recommendations]
\label{corollary:bandits-informative-recos}
%Suppose Assumptions \ref{assn:consistency}, \ref{assumption:parameter-set}, \ref{assumption:noise} hold. Define the (non-negative) primitive
Consider homogeneous agents. Fix any semantic map $\smap$ such that $\delTS>0$ and moreover $\epsTS>0$, where
\begin{align}
    \epsTS := \min_{m\in\msgs,\; j\in\arms\setminus \cbrac{i}}
        \Delta_{i,j}\pbrac{\xsingleton,m},
    \text{ where } i = \menu_Q(\xsingleton,m).
     \label{eqn:primitives-bandits-informative-recos-eps}
\end{align}
%If $\delTS$ (defined in \eqref{eqn:primitives-bandits-informative-recos-delta}) and $\epsTS$ (defined in \eqref{eqn:primitives-bandits-informative-recos-eps}) are positive, and
Further, suppose the warm-up data satisfies
    $\emin \gtrsim \pbrac{R/\epsTS}^2\,\log\pbrac{2/\delTS}$
almost surely.
Then \ourAlg is BIC, and in fact $\eps$-strong-BIC with $\eps = \epsTS/2>0$.
\end{corollary}

%\textbf{Natural menu-consistent semantic mappings \`a la Assumption~\ref{assn:consistency}.}

We next provide several examples.
\begin{enumerate}
    \item The usual direct recommendation message set $\cal{M} = \cal{K}$, together with the semantic mapping
$\smap(\xsingleton, \model) = \argmax_{i\in\cal{K}}\model_i$, with ties broken according to some fixed rule.

\item The "ranking" message set given by the symmetric group over $\cal{K}$, i.e., $\cal{M} = S\pbrac{\cal{K}}$, endowed with the map $\smap(\xsingleton, \model) = \cbrac{\model_{(i)} : i\in \cal{K}}$, with ties broken according to some fixed rule, and where $\model_{(i)}$ is the $\ith{i}$ largest coordinate of $u$.

As a specific example, if $\cal{P}_0$ is Uniform over $\cal{U} = [0,1]^K$, then $\delTS = 1/K!$ and $\epsTS = \Omega\pbrac{1/K}$ (the latter follows using Lemma~3.2 of \cite{sellke2023incentivizing}). Thus, it suffices to collect $\cal{O}\pbrac{K^2\log\pbrac{K!}} = \cal{O}\pbrac{K^3\log K}$ warm-up samples of each arm for \ourAlg to be BIC with rankings as messages.

\item Posterior sampled model is perfectly revealed, i.e, the message set is the model set itself. In this case, $\cal{M} = \MODELS$ and  $\smap: \pubtypes \times \MODELS \mapsto \MODELS$ is the identity map w.r.t. its second input, i.e., $\smap(\xsingleton, \model) = \model\ \forall\ \model\in\MODELS$. For a specific example, consider the case of a finite $\MODELS$.

\end{enumerate}

Of course, all examples implicitly assume a prior $\cal{P}_0$ under which $\epsTS,\delTS$ are strictly positive.

\subsection{Correlated Bayesian priors}
\label{subsection:correlated-bayesian-priors}

This subsection zooms in on the ``problem dimension" of correlated priors. We keep the rest of the problem simple, focusing on homogeneous agents and the trivial semantic map
    $\smap(\xsingleton, \model) = \argmax_{i\in[K]}\model_i$,
with ties broken according to some fixed rule (where $\xsingleton$ is the unique type). Fix the noise parameter to $R=1$.

%In what follows,  dimension $d=K$,  type set $\types = \cbrac{\cal{I}_{K}}$,  model set $\cal{U} = [0,1]^K$, action set $\cal{K} = [K]$, and noise parameter $R=1$.  We consider the standard message set comprising of direct recommendations,  i.e.,  $\cal{M} = [K]$.  The semantic map is given by $\smap(\xsingleton, \model) = \arg\max_{i\in[K]}\model_i$ with ties broken according to some fixed rule, where $\xsingleton$ denotes the unique element of $\types$.

\begin{corollary}[homogeneous agents with correlated priors]
%[\OurAlg is BIC for $K$-armed bandits]
\label{corollary:bandit}
Consider homogeneous agents and semantic map
    $\smap(\xsingleton, \model) = \argmax_{i\in[K]}\model_i$
(up to some consistent tie-breaking). Let
\begin{subequations}
\label{eqn:primitives-standard}
\begin{align}
&\epsTS := \min_{\substack{i,j\in[K]: i \neq j}}{\E_0}\sbrac{\left.  \pbrac{\trueModel_i - \trueModel_j}_+ \right\rvert \smap\pbrac{\xsingleton, \trueModel} = i}, \label{eqn:eps-K-armed} \\
%&\delTS := \min_{i\in[K]}\Prob[0]{\smap\pbrac{\xsingleton, \trueModel} = i}, \\
&\NTS := \frac{C}{\epsTS^2}\log\pbrac{\frac{2}{\delTS}},
\;\text{for some absolute constant $C>0$}.
\end{align}
\end{subequations}
%Suppose Assumptions~\ref{assumption:parameter-set} and \ref{assumption:noise} hold.
If $\epsTS$ and $\delTS$ are positive, and the warm-up data contains at least $\NTS$ samples of each arm almost surely, then \ourAlg is $\eps$-strong-BIC for $\eps = \epsTS/2>0$.
\end{corollary}

%Proof is provided in \S\ref{proof:corollary-bandit}.

%\textbf{Comparison with prior work.}
\begin{remark}
The special case when the prior is independent across arms appeared in \cite{Selke-PoIE-ec21}, with parameters $\epsTS,\,\delTS,\,\NTS$ defined in the same exact way. The analysis in \cite{Selke-PoIE-ec21} relies on the FKG inequality, which is unapplicable for correlated priors. The case of two arms with correlated priors has been treated in \cite{CombiIE-neurips22}, with essentially the same parameters (and a technique that does not appear to extend beyond two arms).
\end{remark}

%Indeed, when the prior is independent across arms, $\epsTS$ is bounded below by $\min_{i,j\in[K] : i\neq j}\E_0\sbrac{\pbrac{\trueModel_i - \trueModel_j}_+}$ due to the FKG inequality (see Lemma A.1 in cited reference), and one recovers from Corollary~\ref{corollary:bandit} the result of \cite{Selke-PoIE-ec21} including identical prior-dependent constants.

%% file: sec-other.tex
In this section, we instantiate our analysis to a few other special cases: sleeping bandits, combinatorial semi-bandits, and linear combinatorial bandits with public types. All three are well-known variants of multi-armed bandits. Throughout, we allow correlated Bayesian prior.

%%%%%%%%%%%%%%%%%%%%%%%%%%%%
\subsection{Sleeping bandits}
\label{subsubsection:sleeping-bandits}

In \emph{sleeping bandits} \citep{sleeping-colt08}, some arms can be ``asleep" in some rounds, \ie not available. Formally, each agent $t$ is endowed with a subset of feasible arms, $\arms_t\subset\arms = [K]$. We posit that this subset is \emph{unobservable} to the principal. This may represent the agent's (hidden) preferences over actions, or ``objective" action availability (\eg a store may run out of a particular item).

We model $\arms_t$ as a private type. Formally, we fix dimension $d=K$,  model set $\MODELS = [0,1]^K$, and identify types as diagonal binary matrices
    $\type_t = \mathtt{diag}(v_t)\in \{0,1\}^{K\times K}$,
where
     $v_t= \rbr{\indi{i\in\arms_t}:\; i\in [K]}$
is the binary representation of $\arms_t$. There's a unique public type, denoted
$\xpub$. For simplicity, the noise parameter is $R=1$. This completes the formal description of the problem.

We define the message set as the set of all rankings (\ie permutations) over arms, denoted $S(\arms)$. We use a convention that for each ranking $\scr{R}\in S(\arms)$,
$\scr{R}(i)$ is the $i$-th ranked arm (so that $\scr{R}(1)$ is the ``best'' arm).
%The message set is rankings, i.e., $\cal{M} = S(\arms)$, for which 
The semantic map $\smap: \pubtypes \times \MODELS \mapsto S(\arms)$, for an input of $\pbrac{\xpub, \model}$, outputs the unique element $\scr{R}$ of $S(\arms)$ satisfying $\model_{\scr{R}(i)} \geq \model_{\scr{R}(j)}\ \forall\ 1 \leq i < j \leq K$, with ties broken according to some fixed rule. Associated with this $\smap$ is the natural menu $\menu_\smap : \types \times S(\arms) \mapsto \arms$ given by $\menu_\smap(x, \scr{R}) = \scr{R}\pbrac{\min\pbrac{i\in [K] : x_{\scr{R}(i)} = \e_{\scr{R}(i)}}}$, where $\e_i$ denotes the $\ith{i}$ standard basis vector of $\R^K$.

%$S(\arms)$ denotes the set of all rankings over arms,
%the symmetric group over $[K]$ (i.e., the set of all permutations of $[K]$.

%We use $\scr{R}$ to refer to an generic element (``ranking'') of $S(\arms)$, with a convention that $\scr{R}(i)$ is the arm occurring at the $\ith{i}$ position (or rank) in $\scr{R}$, so that $\scr{R}(1)$ is the ``best'' arm according to this ranking.

%$x_t = \texttt{diag}\pbrac{a_1(t), ..., a_K(t)}$, where $a_i(t)\in\{0,1\}$ is a \emph{latent} indicator of whether arm $i$ is feasible for agent $t$ or not.  Since the $a_t(t)$'s are latent, we have $\mod{\pubtypes} = 1$. Since nothing about the type sequence $\pbrac{\scr{X}_t : t\in\N}$ is observable to the principal, we have $\mod{\pubtypes} = 1$; let $\xpub$ denote the unique element of this set.

%We will refer to agent $t$'s feasible subset of arms by $\cal{K}_{x_t} \subseteq [K]$.

% \begin{definition}[Best feasible action]
% The best feasible action for an agent with type $X$ (feasible action set $\cal{K}_X\subseteq [K]$) that receives a message $\scr{R}\in S(\arms)$ is given by $A\pbrac{X,\scr{R}} = \scr{R}\pbrac{\min \pbrac{i : \scr{R}(i) \in \cal{K}_X}}$.
% \end{definition}

\begin{corollary}[$K$-armed sleeping bandits]
\label{corollary:sleeping}
Consider the special case of sleeping bandits with the formulation and the semantic map $\smap$ as defined above. Define prior-dependent constants
\begin{align}
&\epsTS := \min_{\scr{R}\in S(\arms)}\min_{i,j\in[K]: i < j}{\E_0}\sbrac{\left.\pbrac{\trueModel_{\scr{R}(i)} - \trueModel_{\scr{R}(j)}}_+ \right\rvert \smap(\xpub, \trueModel) = \scr{R}}, \label{eqn:epsTS-ranking} \\
&\delTS := \min_{\scr{R}\in S(\arms)}\Prob[0]{\smap(\xpub, \trueModel) = \scr{R}}, \label{eqn:delTS-ranking} \\
&\NTS := \frac{C}{\epsTS^2}\log\pbrac{\frac{2}{\delTS}},
\;\text{for some absolute constant $C>0$}.
 \label{eqn:NTS-ranking}
\end{align}
%where $C>0$ is some universal constant.
%Suppose Assumptions~\ref{assumption:parameter-set} and \ref{assumption:noise} hold.
If $\epsTS$ and $\delTS$ are positive, and the warm-up data contains at least $\NTS$ samples of each arm almost surely, then \ourAlg is $\eps$-strong-BIC for $\eps=\epsTS/2>0$.
\end{corollary}

%Proof is provided in \S\ref{proof:corollary-sleeping}.

%An implication of Corollary~\ref{corollary:sleeping} is that \ourAlg inherits the performance of vanilla Thompson Sampling for \emph{observable} feasible arm-subsets after sufficiently many samples of each arm have been observed.

Accordingly, we recover the regret guarantees achieved for Thompson Sampling in sleeping bandits after the warm-start. Indeed, one obtains Bayesian regret of order $K\log T\sqrt{T}$
\citep[Prop. 3]{Russo-MathOR-14}, and also an instance-dependent regret bound
%or \emph{instance-dependent} regret) in sleeping bandits, a bound
of order $\frac{K\log T}{\Delta_{\min}}$, where $\Delta_{\min} := \min_{i,j\in[K]: i \neq j} \left\lvert \trueModel_ i - \trueModel_j \right\rvert$
%and the $\cal{O}(1)$ term depends on $\Delta_{\min}$,   is provided i
\citep{chatterjee2017analysis},
for Bernoulli rewards and independent uniform priors.

%%%%%%%%%%%%%%%%%%%%%%%%%%%%%%%%%%%%%%%%%%%%%%%%%%%%%%%%%%%%%%%%%%%%%%%%%%%%%%%%%%%%%%%%

\subsection{Combinatorial semi-bandits}
\label{sec:combi-bandits}

In combinatorial semi-bandits 
\citep[\eg][]{Chen-icml13,Kveton-aistats15,MatroidBandits-uai14},
arms correspond to subsets of ``atoms": there is a set of $d$ ``atoms", denoted $[d]$, and the action set is $\arms \subseteq 2^{[d]}$. Each atom generates a reward, which is observed.  This problem can be ``embedded" as a special case of $d$-dimensional linear contextual bandits with semi-bandit feedback: each arm $\arm\subset [d]$ is identified with the indicator vector $x_{\arm} = \rbr{\indi{j\in A}:\,j\in [d] }$, and there is a unique public context which comprises these indicator vectors.

%Each arm $\arm\in\arms$  generates a mean reward of $\nu\pbrac{\arm; \mu} := \sum_{j\in \arm}\mu_j$.

%In this setting,  $\MODELS = [0,1]^d$, there are $d$ atoms (denoted by $[d]$) and the set of arms is $\arms \subseteq 2^{[d]}$ with $K = |\arms|$.  Each arm is some subset of $[d]$, i.e., $\arm\subseteq [d]$, and

%generates a mean reward of $\nu\pbrac{\arm; \mu} := \sum_{j\in \arm}\mu_j$.  The semi-bandit feedback observed upon playing arm $\arm$ in round $t\in\N$ is given by $\cbrac{r_{j,t} = \mu_j + \xi_{j,t} : j\in \arm}$, where the $d$-dimensional noise process $\pbrac{\xi_t : t=1,2,\dots}$ satisfies Assumption~\ref{assumption:noise}.

%the usual message space of direct recommendations, i.e., $\cal{M} = \arms$.  Since $\mod{\cal{X}} = 1$, the semantic map is given by $\smap\pbrac{\xsingleton, \model} = \arg\max_{\arm\in\arms}\nu\pbrac{\arm; \model}$, with ties broken according to some fixed rule, where $\xsingleton$ denotes the unique element of $\cal{X}$.

\begin{corollary}[combinatorial semi-bandits]
\label{corollary:combinatorial}
Consider the special case of combinatorial semi-bandits, as defined above. Consider direct recommendations, with message set $\msgs = \arms$ and semantic map $\smap = \menu^*$. Define prior-dependent constants
\begin{align*}
&\epsTS := \min_{\substack{\arm,\armprime\in\arms \\ \arm \neq \armprime}}{\E_0}\sbrac{\left.\pbrac{\nu\pbrac{\arm; \trueModel} - \nu\pbrac{\armprime; \trueModel}}_+ \right\rvert \smap(\xsingleton, \trueModel) = \arm}, \\
&\delTS := \min_{\arm\in\arms}\Prob[0]{\smap(\xsingleton, \trueModel) = \arm}, \\
&\NTS = \frac{C\s^2}{\epsTS^2}\log\pbrac{\frac{2}{\delTS}}
\;\text{for some absolute constant $C>0$}.
\end{align*}
%Suppose that Assumptions~\ref{assumption:parameter-set} and \ref{assumption:noise} hold. 
If $\epsTS$ and $\delTS$ are positive and the warm-up data contains at least $\NTS$ samples of each atom $j\in[d]$ almost surely, then Thompson Sampling is $\eps$-strong-BIC, $\eps=\epsTS/2>0$.
\end{corollary}

% Proof is provided in \S\ref{proof:corollary-combinatorial}.
%\textbf{Comparison with prior work.} 

Note that the sparsity parameter $\s$ from Assumption~\ref{assumption:parameter-set} is upper-bounded as $\s \leq d$.

\begin{remark}
The special case of the prior that is independent across the atoms
%A version of Corollary~\ref{corollary:combinatorial} for the special cases of (i) a product-form (independent) Bayesian prior, and (ii) correlated prior with $d=2$, 
appeared in \cite{CombiIE-neurips22}, with essentially the same parameters (upper-bounding $\s$ with $d$). However, their technique, based on the mixed-monotonicity Harris inequality, does not appear to extend to general priors.
%. Indeed, when the prior is independent across atoms, $\min_{\substack{\arm,\armprime\in\arms: \arm \neq \armprime}}{\E_0}\sbrac{\pbrac{\nu\pbrac{\arm; \trueModel} - \nu\pbrac{\armprime; \trueModel}}_+}$ lower bounds $\epsTS$ due to the mixed-monotonicity Harris inequality (see Theorem C.2 in cited reference), and one recovers from Corollary~\ref{corollary:combinatorial} the result of \cite{hu2022incentivizing} including identical prior-dependent constants after using $d$ to upper bound the sparsity $\s$.
\end{remark}

%%%%%%%%%%%%%%%%%%%%%%%%%%%%%%%%%%%%%%%%%%%%%%%%%%%%%%%%%%%%%%%%%%%%%%%%%%%%%%%%%%%%%%%%

\subsection{Linear bandits with public types}
\label{subsection:CBpublic}

We finally turn our attention to consider linear contextual bandits with public types. We consider direct recommendations, with $\MODELS = \arms$ and semantic map $\smap=\menu^*$. We recover a result from \citep{sellke2023incentivizing} in part (1) below, and then extend it to allow for an adaptive warm-start in part (2). 

% the same same setting as \S\ref{subsection:CBprivate}, now with observable type sequences $\pbrac{\scr{X}_t: t=1,2,...}$, i.e., $\pubtypes = \cal{X}$. 

%. The semantic map is given by $\smap(x,\model) = \arg\max_{i\in\arms}x_i \cdot \model$, with ties broken according to some fixed rule.

We restate below the primitives from \eqref{eqn:primitives} under this simplified semantic mapping.
\begin{subequations}
\label{eqn:primitives-public}
\begin{align}
%&\menuset_{\smap}(x,k) :=
%    \bigcup_{\text{models }\model \in \MODELS:\;\; \smap(x, \model) = k}\quad
%    \argmax_{\text{arms } i\in\arms}\; x_i\cdot\model,   \\
&\Delta_{i,j}\pbrac{x,k} := \E_0
    \sbr{ \pbrac{x_i - x_j}\cdot \trueModel \mid \smap\pbrac{x, \trueModel} = k },  \\
&\delTS := \inf_{x\in\cal{X}}\min_{k\in\arms}\Prob[0]{\smap(x, \trueModel) = k}.\label{eqn:primitives-public-delta}
\end{align}
\end{subequations}
Here,  type $x\in\cal{X}$ and arms $i,j,k\in\arms$.

\begin{assumption}[$\alpha$-separated types on the norm ball]
\label{assumption:alpha-separation}
  Each $x\in\cal{X}$ satisfies $\norm{x_{i}}_2 = 1\ \forall\ i\in\arms$, and $\norm{x_{i} - x_{j}}_2 \geq \alpha$ for some $\alpha\in(0,1)$ $\forall$ distinct $i,j\in\arms$.
\end{assumption}

\begin{assumption}[Uniform prior over the norm ball]
\label{assumptipon:uniform-prior-CB}
    The prior $\cal{P}_0$ is Uniform over the model set
    $\cal{U} = \cbrac{u \in\R^d : \norm{u}_2 \leq 1}$.
\end{assumption}

Assumptions~\ref{assumption:alpha-separation}, \ref{assumptipon:uniform-prior-CB}
ensure $\Delta_{i,j}(x,i) \gtrsim \alpha^2/d$ for any two distinct arms $i,j$, and $\delTS \geq \pbrac{\alpha/4}^d$.

\begin{corollary}
%[\OurAlg is BIC for linear contextual bandits]
\label{corollary:CB-public}
Suppose
%~\ref{assumption:parameter-set}, \ref{assumption:noise}, 
 Assumptions~\ref{assumption:alpha-separation} and \ref{assumptipon:uniform-prior-CB} hold. Suppose the spectral diversity of the warm-up data satisfies 
    $\emin \gtrsim \pbrac{D/\alpha^2}\log\pbrac{8/\alpha}$ 
 almost surely, where $D$ is specified below. Further, suppose one of the following holds:
\begin{enumerate}

\item Warm-up data is non-adaptive, has bandit feedback, and $D = d^3R^2$.
\item Warm-up data has bandit feedback, and $D = d^4\pbrac{R + 1}^2\log\pbrac{T+3}$.

\end{enumerate}

Then, Thompson Sampling is $\eps$-strong-BIC with $\eps = \pbrac{\alpha^2/d}>0$.

\end{corollary}

%Proof is provided in \S\ref{proof:corollary-public}.

%\textbf{Comparison with prior work.} 
%Corollary~\ref{corollary:CB-public} strengthens Theorem~3.5 of \cite{sellke2023incentivizing} to allow for an adaptive warm-start.
%%%%%%%%%%

%% file: IE-via-others.tex
\section{Incentivized exploration via other native algorithms}
\label{section:other-algorithms}

While the discussion so far has focused only on \ourAlg, it is noteworthy that the same proof techniques, in essence, can be used to analyze and establish results for other online learning algorithms as well. This section is dedicated to providing such results for two well-known algorithms; Least-Squares-Greedy for linear bandits \cite{bastani2021mostly,raghavan2023greedy}, and UCB (and Frequentist-Greedy as a special case) for $K$-armed bandits \cite{bandits-ucb1}.

\subsection{\OLS algorithm for linear bandits with private types}

\OLS generates the model $\model_t$ in round $t$ using the regularized least squares estimator $\hat{\model}_t := \pbrac{\cal{I}_d + \hat{\Sigma}_{t-1}}^{-1}\sum_{s\in[t-1]}\type_{\arm_s,s} y_s$ of $\trueModel$. Regularized least squares is a well-studied estimator for linear models \cite{Csaba-nips11} that also serves for us as a natural baseline to benchmark the guarantees of \ourAlg.

We consider the linear bandit setting with private types, i.e., $\mod{\pubtypes} = 1$, and denote by $\xpub$ the unique element of $\pubtypes$.

\begin{assumption}[Perfect hypercube-covering of model set $\cal{U}$]
\label{assumption:hypercube-covering}
There exists a finite collection of hypercubes ($\ell_\infty$ balls) each of radius $\varepsilon$, given by $\bb{U} = \cbrac{U_m : m\in \cal{M}}$ where $\cal{M} \subseteq \N$, such that (i) $U_m \cap U_n = \phi\ \forall$ distinct $m,n\in \cal{M}$, (ii) $U_m \subseteq \cal{U}\ \forall\ m\in \cal{M}$, and (iii) $\bigcup_{m\in \cal{M}}U_m = \cal{U}$.
\end{assumption}

Assumption~\ref{assumption:hypercube-covering} is generically satisfied if the model set $\cal{U}$ is some $\ell_\infty$ ball itself.

\textbf{Message set, semantic map, and primitives.} We fix $\cal{M}$ as the message set and consider the mapping $\scr{Q} : \pubtypes \times \cal{U} \mapsto \cal{M}$ given by 
\begin{align}
    \smap\pbrac{\xpub, \model} := \sum_{m\in \cal{M}}m\indi{\model\in U_m}.
\end{align}

% We restate below the relevant primitives from \eqref{eqn:primitives-private} for the semantic map defined above.
% \begin{align}
% &\menuset_{\smap}(x,m) :=
%     \bigcup_{\text{models }\model \in \MODELS:\;\; \smap(\xpub, \model) = m}\quad
%     \argmax_{\text{arms } i\in\cal{K}}\; x_i \cdot \model, \label{eqn:plausible-best-arms-OLS} \\
% &\Delta_{i,j}\pbrac{x,m} := \E_0
%     \sbr{\pbrac{x_i - x_j}\cdot\trueModel \mid \smap\pbrac{\xpub, \trueModel} = m } \\
% &\delTS := \inf_{x\in\cal{X}}\min_{m\in\cal{M}}\Prob[0]{\smap(\xpub, \trueModel) = m}. \label{eqn:primitives-OLS-delta}
% \end{align}

In addition, our analysis also depends on the following prior-dependent constant
\begin{align}
    \eta := \frac{\delTS}{\pbrac{2\varepsilon}^d\sup f}, \label{eqn:primitives-OLS-eta}
\end{align}
where $f: \cal{U} \mapsto \bb{R}_+$ is the density associated with prior $\cal{P}_0$. Note that $\delTS$ is the smallest prior probability of any subset $U_m$ of the model set $\MODELS$,  and $\eta$ is a lower bound on the ratio of $\delTS$ to the largest prior probability of any $U_m$. E.g., if $\cal{P}_0$ is Uniform over $\cal{U}$, i.e., $f(u) = 1/\vol{\cal{U}}\ \forall\ u\in\cal{U}$, then $\delTS = \pbrac{2\varepsilon}^d / \vol{\cal{U}}$ and $\eta=1$. 

\begin{theorem}[\OLS needs many warm-up samples to achieve BIC]
\label{thm:BIC-OLS}

Suppose Assumption~\ref{assumption:hypercube-covering} holds, and that $\delTS$ and $\eta$ are positive. Further suppose that $\smap$ is $\varepsilon$-menu-consistent for some fixed $\varepsilon>0$, and that the spectral diversity of the warm-up data satisfies almost surely
\begin{align}
\emin \gtrsim \frac{d^3C_{\cal{U}}^4C_{\cal{X}}^4\pbrac{RC_{\cal{X}} + C_{\cal{U}}}^2}{\varepsilon^4\eta^2}\log\pbrac{C_{\cal{X}}^2T + 3}\log\pbrac{\frac{16eC_{\cal{U}}C_{\cal{X}}}{\varepsilon\delTS}}.
\end{align}
Then, \ols is $(\varepsilon\delTS/4)$-BIC.
\end{theorem}

Proof is deferred to \S\ref{proof:BIC-OLS}. 

\begin{remark}
In comparison, \ourAlg only needs $\emin\gtrsim \Omega\pbrac{1/\varepsilon^2}$ to achieve this guarantee, ceteris paribus (cf.  Corollary~\ref{corollary:CB-private}). Moreover, it also does not rely on the $\eta>0$ assumption.
\end{remark}

While Theorem~\ref{thm:BIC-OLS} (in conjunction with Corollary~\ref{corollary:CB-private}) is insufficient for one to conclude that \ols is an inferior choice for incentivized exploration via-\`a-vis \ourAlg, the fact that a gap of $\Omega\pbrac{1/\varepsilon^2}$ exists between the two lower bounds, and that they have been derived under identical conditions and using nearly the same high-level approach and technical machinery, does make the case for \ourAlg stronger.

\subsection{UCB and Frequentist-Greedy algorithms for K-armed bandits}

We will work with the standard model set $\cal{U} = [0,1]^K$, and assume a noise parameter of $R=1$. Let $N_i(t)$ denote the number of times arm~$i$ is played up to (excluding) round $t\in\N$, and let $\hat{\model}_{i,t}$ be its empirical mean reward just prior to the start of round $t$. Define $\hat{\model}_t := \pbrac{\hat{\model}_{i,t} : i\in[K]}$. Fix some $\rho>0$ and define $\hat{\model}^{\texttt{UCB}}_{i,t} := \hat{\model}_{i,t} + \sqrt{\frac{\rho\log\pbrac{t-1}}{N_i(t)}}$ for $i\in[K]$,  and $\hat{\model}^{\texttt{UCB}}_{t} := \pbrac{\hat{\model}^{\texttt{UCB}}_{i,t} : i\in[K]}$. 

Our analysis requires the prior distribution to satisfy a mild regularity condition of the form introduced in \cite{tsybakov2004optimal}.

\begin{assumption}[$\alpha$-margin condition on prior]
\label{assumption:alpha-margin}
There exists some $\alpha>0$ such that the prior $\cal{P}_0$ satisfies with a universal constant $C>0$ the following condition:
\begin{align*}
\Prob[\trueModel\sim\cal{P}_0]{\mod{\trueModel_i - \trueModel_j} \leq \varepsilon} \leq C\varepsilon^\alpha\ \ \ \ \ \ \ \ \ \ \ \forall\ \text{distinct}\ i,j\in[K],\ \forall\ \varepsilon>0.
\end{align*}
\end{assumption}

\textbf{Message set and mechanism.} We consider the usual message set comprising of direct recommendations, i.e., $\cal{M} = [K]$. The semantic map is given by $\smap(\xsingleton, \model) = \arg\max_{i\in[K]}\model_i$,  with ties broken according to some fixed rule, and where $\xsingleton = \cal{I}_d$.

\UCB generates the model $\model_t$ in round $t$ using the UCB algorithm \cite{bandits-ucb1}, which reduces to the Greedy algorithm when when $\rho=0$.

\begin{theorem}[\UCB needs many warm-up samples to achieve BIC]
\label{thm:BIC-UCB}
Suppose Assumption~\ref{assumption:alpha-margin} holds, and that $\epsTS$ and $\delTS$ are positive. For some large enough universal constant $C>0$, define the primitives
\begin{align}
    &\epsUCB := \pbrac{\frac{\epsTS\delTS}{CK}}^{\frac{1}{\alpha}} \label{eqn:epsUCB} \\
    &\NUCB := \pbrac{\frac{\alpha + 2}{\epsUCB^2}}\log\pbrac{\frac{1}{\epsUCB}} + \frac{\rho\log T}{\epsUCB^2}. \label{eqn:NUCB}
\end{align}

If the warm-up data contains at least $\NUCB$ samples of each arm $i\in[K]$ almost surely, then UCB is $\eps$-strong-BIC with $\eps = K\epsUCB^\alpha/2$.

\end{theorem}

The proof is provided in \S\ref{proof:BIC-UCB}. 

\begin{remark}
In comparison, Thompson Sampling only needs approximately $\frac{1}{\epsTS^2}\log\pbrac{\frac{2}{\delTS}}$ samples of each arm to, in fact, achieve a stronger guarantee (cf.  Corollary~\ref{corollary:bandit}). Moreover, it also does not rely on Assumption~\ref{assumption:alpha-margin} to achieve BIC.
\end{remark}

%% file: standard_definitions.tex
\newpage
\section{Standard definitions, facts, and results used in analyses}
\label{section:facts}

\begin{definition}[Sparsity]
\label{definition:sparsity}
    A vector $\nu\in\R^d$ is said to be ``$\s$-sparse'' iff $\norm{\nu}_0 \leq \s$. Here, $\norm{\nu}_0$ is the ``$\ell_0$'' norm of $\nu$, which is the number of non-zero coordinates of $\nu$.
\end{definition}

\begin{definition}[Sub-Gaussianity]
\label{definition:subG}
We say that a real-valued random variable $W$ is sub-Gaussian with variance proxy $R^2$ (denoted as $W\sim\subG\pbrac{R^2}$) iff $\E W = 0$ and $\E\sbrac{\exp\pbrac{\theta W}} \leq \exp\pbrac{{R^2\theta^2}/{2}}\ \forall\ \theta\in\R$.
\end{definition}

\begin{fact}[Sub-Gaussian tail bound]
\label{fact:subG-tail}
    If $W\sim\subG\pbrac{R^2}$, then $\max\cbrac{\Prob{W \geq \theta}, \Prob{W \leq -\theta}} \leq \exp\pbrac{-\theta^2/\pbrac{2R^2}}$ for any $\theta > 0$.
\end{fact}

\begin{fact}[Sufficient condition for sub-Gaussianity $\sbrac{\text{Lemma~1 of \cite{shamir2011variant}}}$]
\label{fact:shamir}
    If a mean-zero random variable $W$ satisfies one of the two equivalent conditions below with some constants $a,b>0$, then $W\sim\subG\pbrac{Ca/b}$ for some universal constant $C>0$.
\begin{enumerate}
    \item $\max\cbrac{\Prob{W \geq \theta}, \Prob{W \leq -\theta}} \leq a\exp\pbrac{-b\theta^2}\ \forall\ \theta>0$.
    \item $\max\cbrac{\Prob{W \geq \sqrt{\pbrac{1/b}\log\pbrac{a/\delta}}}, \Prob{W \leq -\sqrt{\pbrac{1/b}\log\pbrac{a/\delta}}}} \leq \delta\ \forall\ \delta>0$.
\end{enumerate}
\end{fact}

\begin{fact}[Gaussian tail bound]
\label{fact:Gaussian-tail}
If $W$ is a standard Gaussian random variable and $\phi(\cdot)$ denotes its density, then $\Prob{W \geq \theta} \leq \frac{\phi(\theta)}{\theta}\ \ \forall\ \theta > 0$.
\end{fact}

\begin{assumption}[Conditional $R$-sub-Gaussian Noise Process]
\label{assumption:noise}
Let $\cal{F}_t$ be the sigma-algebra generated by all of principal's observations during all rounds up to and including round $t$.
%    $\cal{F}_{t-1}:= \sigma\left( {\type_s}, \arm_s,Y_s: s\in[t-1]\right)$. 
Let ${\xi}_t := \pbrac{\xi_{j,t} : j\in[d]}$. The zero-mean noise process $\rbr{ {\xi}_t \in\R^d : t = 1,2,... }$ satisfies the following
    \begin{align*}
        \left. w\cdot{\xi}_t \;\right\rvert\sigma\rbr{ {\type_t}, \arm_t, \cal{F}_{t-1}, \trueModel} \sim \subG\rbr{\norm{w}_2^2\; R^2}
    \quad\text{for each round $t\in\N$ and any $w\in\sigma\rbr{ {\type_t}, \arm_t, \cal{F}_{t-1}, \trueModel}$.}
    \end{align*}    
\end{assumption}

Assumption~\ref{assumption:noise} subsumes the canonical i.i.d. noise specification as well as allows for correlated noise processes (across arms/atoms and time).

\begin{fact}[H{\"o}lder's inequality]
\label{fact:holder}
Let $\norm{\cdot}:\R^d \mapsto \R$ be some norm.  Define its dual norm $\norm{\cdot}_\ast : \R^d \mapsto \R$ as $\norm{y}_{\ast} := \sup_{w\in\R^d}\pbrac{w\cdot y : \norm{w} \leq 1}$. Then, $w\cdot y \leq \norm{w}\norm{y}_\ast$ holds for any $w,y\in\R^d$.
\end{fact}

\begin{fact}[Weighted/Matrix norms]
\label{fact:matrix}
Let $M\in\R^{d\times d}$ be some positive definite matrix. Define $\norm{\cdot}_M : \R^d \mapsto\R$ as $\norm{y}_M := \sqrt{y\cdot My}$. Then, $\norm{\cdot}_M$ is a norm and $\norm{\cdot}_{M^{-1}}$ is its dual norm. Further,  the following holds for any $y\in\R^d$
\begin{align*}
\frac{\norm{y}_2}{\sqrt{\lambda_{\max}\pbrac{M}}} = \norm{y}_2\sqrt{\lambda_{\min}\pbrac{M^{-1}}} \leq \norm{y}_{M^{-1}} \leq \norm{y}_2\sqrt{\lambda_{\max}\pbrac{M^{-1}}} = \frac{\norm{y}_2}{\sqrt{\lambda_{\min}\pbrac{M}}}.
\end{align*}
\end{fact}

\begin{fact}[Courant-Fischer Theorem]
\label{fact:courant-fischer}
    Let $M\in\R^{d\times d}$ be a symmetric matrix with maximum and minimum eigenvalues $\lambda_{\max}\rbr{M}$ and $\lambda_{\min}\rbr{M}$ respectively. Then,
    \begin{align*}
    \lambda_{\max}\rbr{M} = \max_{\substack{\omega\in\R^d \\ \norm{\omega}_2 =1}}\omega \cdot M\omega \quad \quad \text{and} \quad \quad\lambda_{\min}\rbr{M} = \min_{\substack{\omega\in\R^d \\ \norm{\omega}_2 =1}}\omega \cdot M\omega.
    \end{align*}
\end{fact}

\begin{fact}[Minimum eigenvalue function $\lambda_{\min}(\cdot)$ is concave over positive semi-definite matrices]
\label{fact:min-eigenvalue}
If $A$ and $B$ are positive semi-definite matrices, then $\lambda_{\min}\pbrac{A+B} \geq \lambda_{\min}\pbrac{A} + \lambda_{\min}\pbrac{B}$.
\end{fact}

\begin{definition}[$\eps$-cover]
\label{definition:eps-cover}
    An $\eps$-cover of a set $X \subseteq \R^d$ w.r.t. norm $\norm{\cdot}$ is a subset $\hat{X}\subset X$ such that for each $\omega \in X$, there exists some $\hat{\omega}\in\hat{X}$ satisfying $\norm{\omega - \hat{\omega}} \leq \eps$.
\end{definition}

\begin{definition}[Voronoi partition]
\label{definition:voronoi}
The Voronoi partition of set $X \subseteq \R^d$ w.r.t. norm $\norm{\cdot}$ and collection of points (aka centers or seeds) $\cbrac{\nu_1, \dots, \nu_N} \subseteq X$ is given by $\cbrac{\scr{P}_i : i\in [N]}$, where $\scr{P}_i := \cbrac{\omega \in X : \norm{\omega - \nu_i} \leq \norm{\omega - \nu_j}\ \forall\ j\in[K]}$, with points on the boundary assigned to exactly one $\scr{P}_i$ according to a fixed rule. 
\end{definition}

It follows that $\cbrac{\scr{P}_i : i\in [N]}$ satisfies (i) $\scr{P}_i \cap \scr{P}_j = \phi\ \forall\ i\neq j$; and (ii) $\cup_{i\in[N]}\scr{P}_i = X$.

\begin{definition}[$\rho$-granular semantic map]
\label{definition:granular}
Semantic map $\smap$ is called \emph{$\rho$-granular}, for some $\rho \geq 0$, if
    $\norm{\model_1 - \model_2}_2 \leq \rho$
for any two models $\model_1,\model_2\in\MODELS$ such that $\smap(\model_1)=\smap(\model_2)$.
\end{definition}

The next result follows easily from Theorem~2 of \cite{Csaba-nips11}.

\begin{lemma}[Theorem~2 of \cite{Csaba-nips11}]
\label{lemma:OLS-abbasi}
Suppose Assumptions~\ref{assumption:parameter-set} and \ref{assumption:noise} hold.  Fix $t\in\N$ and define for $\cal{T} \subseteq [t-1]$ the regularized OLS estimator $\model_t := \pbrac{\cal{I}_d + \hat{\Sigma}_{\cal{T}}}^{-1}\sum_{s\in\cal{T}}x_{A_s, s}y_s$. Then, the following holds for any $\delta>0$:
\begin{align*}
\Prob{\left. \norm{ \model_t - \trueModel  }_{I_d + \hat{\Sigma}_{\cal{T}}} \geq \pbrac{RC_{\cal{X}} + C_{\cal{U}}}\sqrt{2d\log\pbrac{C_{\cal{X}}^2t + 3}\log\pbrac{\frac{e}{\delta}}}  \right\rvert \trueModel } \leq \delta.
\end{align*}

\end{lemma}

\textit{Proof of Lemma~\ref{lemma:OLS-abbasi}.} We have $y_t = x_{A_t, t}\cdot \trueModel + x_{A_t, t}\cdot \xi_t$. Define $\eta_t := x_{A_t, t}\cdot \xi_t$, and $F_t := \sigma\pbrac{\cbrac{\scr{X}_s, A_s : s\in[t+1]}, \cbrac{y_s : s\in[t]}, \trueModel}$ for $t\in\N$. Then, $\eta_t \in F_t$ and by Assumption~\ref{assumption:noise}, $\eta_t | F_{t-1} \sim \subG\pbrac{C_{\cal{X}}^2R^2}$ for all $t\in\N$. Theorem~2 of \cite{Csaba-nips11} then implies that for any subset of rounds $\cal{T}\subseteq [t-1]$ and $\delta>0$,
\begin{align}
    \Prob{\left. \norm{ \model_t - \trueModel  }_{I_d + \hat{\Sigma}_{\cal{T}}} \geq    RC_{\cal{X}}\sqrt{d\log\pbrac{\frac{C_{\cal{X}}^2t+1}{\delta}}} + C_{\cal{U}}    \right\rvert \trueModel} \leq \delta. \label{eqn:abbasi-OLS1}
\end{align}
Now observe that 
\begin{align}
    RC_{\cal{X}}\sqrt{d\log\pbrac{\frac{C_{\cal{X}}^2t+1}{\delta}}} + C_{\cal{U}} &\leq \pbrac{RC_{\cal{X}} + C_{\cal{U}}}\sqrt{d\log\pbrac{\frac{C_{\cal{X}}^2t+3}{\delta}}} \notag \\
    &= \pbrac{RC_{\cal{X}} + C_{\cal{U}}}\sqrt{d\log\pbrac{\frac{1}{\delta}} + d\log\pbrac{C_{\cal{X}}^2t+3}} \notag \\
    &\leq \pbrac{RC_{\cal{X}} + C_{\cal{U}}}\sqrt{d\log\pbrac{\frac{e}{\delta}} + d\log\pbrac{C_{\cal{X}}^2t+3}} \notag \\
    &\leq \pbrac{RC_{\cal{X}} + C_{\cal{U}}}\sqrt{2d\log\pbrac{C_{\cal{X}}^2t+3}\log\pbrac{\frac{e}{\delta}}}. \label{eqn:abbasi-OLS2}
\end{align}
Combining \eqref{eqn:abbasi-OLS1} and \eqref{eqn:abbasi-OLS2} establishes the stated assertion. \qed

The following technical lemma is adapted from Lemma~2.1 in \cite{sellke2023incentivizing}. 

\begin{lemma}[Bayesian-Chernoff bound]
\label{lemma:Bayesian-Chernoff}
Let $\left( \Omega, \mathfrak{F}, \left\lbrace F_t\right\rbrace_{t=0}^\infty, \bb{P} \right)$ be a filtered probability space. Let $\nu\in V\subseteq \bb{R}^d$ be a random vector defined on this space.  Fix $t\in\N$.  Suppose $\hat{\nu}_t\in\ V$ is an $F_t$-measurable estimator for $\nu$ that satisfies for some deterministic $\varepsilon, \delta>0$ and $b\in\R^d$ the concentration inequality
\begin{align}
\Prob{ \left. \left\lvert b\cdot \pbrac{\hat{\nu}_t - \nu} \right\rvert \geq \varepsilon \right\rvert \nu} \leq \delta\ \ \ \ \forall\ \nu\in V. \label{eqn:BC-ref}
\end{align}
If $\nu$ is distributed according to a prior distribution $\cal{P}_0$ and $\tilde{\nu}_t$ is sample from its posterior distribution given $F_t$, then the following holds:
\begin{align*}
\bb{P}_{\nu\sim\cal{P}_0,\ \tilde{\nu}_t\sim \nu | F_t}\left( \left\lvert b\cdot \pbrac{\tilde{\nu}_t - \nu} \right\rvert \geqslant 2\varepsilon \right) \leqslant 2\delta.
\end{align*}
\end{lemma}

\noindent\textit{Proof of Lemma~\ref{lemma:Bayesian-Chernoff}:} Note that the tuples $\left( F_t, \nu \right)$ and $\left( F_t, \tilde{\nu}_t\right)$ are identically distributed. Since $\hat{\nu}_t$ is $F_t$-measurable, it follows that $\left( \hat{\nu}_t, \nu\right)$ and $\left(\hat{\nu}_t,  \tilde{\nu}_t \right)$ are identically distributed as well; both tuples therefore satisfy \eqref{eqn:BC-ref}. The claim now follows by the triangle inequality. \hfill $\square$

The following result for random matrices is adapted from Proposition~1 of \cite{li2017provably}.

\begin{lemma}[Proposition~1, \cite{li2017provably}]  
\label{fact:vershynin}
Suppose $\hat{\Sigma}_{[t]} := \textstyle \sum_{s\in[t]}{X_s}\otimes{X_s}$, where $({{X}_s}\in\bb{R}^d : s = 1,2,...)$ is an i.i.d. sequence of random vectors with distribution supported in the $\ell_2$ norm ball $\cbr{\nu\in\R^d : \norm{\nu}_2 \leq 1}$. Then, for any $t \geqslant 1$ and $\delta>0$, $\lambda_{\min} \left( \hat{\Sigma}_{[t]} \right) \geqslant \lambda_0 t - C_1\sqrt{dt} - C_2\sqrt{t\log(1/\delta)}$ holds with probability at least $1-\delta$, where $\lambda_0 := \lambda_{\min} \left( \E\sbr{X_1 \otimes X_1} \right)$ and $C_1,C_2>0$ are universal constants. 
\end{lemma}

\begin{lemma}
\label{lemma:eigen_min}
    Let $X$ be a random matrix in $\cal{X} \subseteq \R^{K \times d}$ and $\arm$ be a random variable taking values in $[K]$. If their joint distribution $F$ satisfies (i) $\lambda_{\min}\rbr{\E\sbr{\frac{1}{K}\sum_{i\in[K]}X_i \otimes X_i}} \geq \lambda_0$ for some fixed $\lambda_0>0$, and (ii) $\Pr[\arm = i \mid X = x] \geq \varepsilon/K$ for some fixed $\varepsilon>0$ and all $i\in[K]$, $x\in\cal{X}$, then one has that $\lambda_{\min}\rbr{\E\sbr{X_{\arm} \otimes X_{\arm}}} \geq \varepsilon\lambda_0$.
\end{lemma}

\textit{Proof of Lemma~\ref{lemma:eigen_min}.} Denote the marginal distribution of $X$ by $D$.  Then,
\begin{align}
    \lambda_{\min}\rbr{\E_{(X,\arm)\sim F}\sbr{X_{\arm} \otimes X_{\arm} }} &= \lambda_{\min}\rbr{\E_{(X,\arm)\sim F}\sbr{\sum_{i\in[K]}X_i \otimes X_i \indi{\arm=i}}} \notag \\
    &= \lambda_{\min}\rbr{ \sum_{i\in[K]} \E_{(X,\arm)\sim F}\sbr{X_i \otimes X_i \indi{\arm=i}}} \notag \\
    &\underset{\star}{=} \min_{\substack{\omega\in\R^d \\ \norm{\omega}_2 = 1}} \omega \cdot \rbr{ \sum_{i\in[K]} \E_{(X,\arm)\sim F}\sbr{X_i \otimes X_i \indi{\arm=i}}}\omega \notag \\
    &= \min_{\substack{\omega\in\R^d \\ \norm{\omega}_2 = 1}}\sum_{i\in[K]} \E_{(X,\arm)\sim F}\sbr{\omega \cdot X_i \otimes X_i \omega \indi{\arm=i}} \notag \\
    &= \min_{\substack{\omega\in\R^d \\ \norm{\omega}_2 = 1}}\sum_{i\in[K]} \E_{\substack{X\sim D}}\sbr{\omega \cdot X_i \otimes X_i \omega \Pr[\arm = i \mid X] }\notag \\
    &\underset{\dag}{\geq} \varepsilon\min_{\substack{\omega\in\R^d \\ \norm{\omega}_2 = 1}}\sum_{i\in[K]} \E_{\substack{X\sim D}}\sbr{\frac{1}{K}\omega \cdot X_i \otimes X_i \omega }\notag \\
    &\underset{\ast}{=} \varepsilon\min_{\substack{\omega\in\R^d \\ \norm{\omega}_2 = 1}} \omega \cdot \rbr{\E_{X\sim D}\sbr{\frac{1}{K}\sum_{i\in[K]}X_i \otimes X_i}} \omega \notag \\
    &= \varepsilon\lambda_{\min}\rbr{\E_{X\sim D}\sbr{\frac{1}{K}\sum_{i\in[K]}X_i \otimes X_i}} \notag \\
    &= \varepsilon\lambda_0, \notag
\end{align}
where $(\star)$ and $(\ast)$ follow using Fact~\ref{fact:courant-fischer}, and $(\dag)$ holds since the outer-product is positive semi-definite. \qed

\begin{lemma}[Growth-rate of minimum eigenvalue during near-uniform exploration]
\label{lemma:near-uniform-exploration}
      Suppose that the sequence $\cbr{\rbr{\type_t, \arm_t} : t = 1,2,...}$ is i.i.d. in time, and that $\rbr{\type_1, \arm_1}$ is distributed according to $F$ satisfying 
    \begin{enumerate}
        \item   $\lambda_{\min}\rbr{\E\sbr{\frac{1}{K}\sum_{i\in[K]}\type_{i,1} \otimes \type_{i,1}}} \geq \lambda_0 > 0$.

        \item $\Pr[\arm_1 = i \mid \type_1 = x] \geq \varepsilon/K$ for some fixed $\varepsilon>0$ and all $i\in[K]$, $x\in\cal{X}$.
    \end{enumerate}

     Let $\hat{\Sigma}_{[t]} := \textstyle \sum_{s\in[t]}{\type_{\arm_s,s}}\otimes{\type_{\arm_s,s}}$. Then, the following holds with probability at least $1-\delta$ for any $t\in\N$ and $\delta>0$:
     $$\lambda_{\min} \left( \hat{\Sigma}_{[t]} \right) \geqslant \varepsilon\lambda_0 t - C_1\sqrt{dt} - C_2\sqrt{t\log(1/\delta)},$$ 
     where $C_1,C_2$ are positive universal constants. 
\end{lemma}

\textit{Proof of Lemma~\ref{lemma:near-uniform-exploration}.} It follows from Lemma~\ref{fact:vershynin} that $\lambda_{\min} \left( \hat{\Sigma}_{[t]} \right) \geqslant \lambda^\prime t - C_1\sqrt{dt} - C_2\sqrt{t\log(1/\delta)}$, where $\lambda^\prime$ is the smallest eigenvalue of $\E_{(\type_1,\arm_1)\sim F}\sbr{\type_{\arm_1,1} \otimes \type_{\arm_1,1} }$. Next, it follows using Lemma~\ref{lemma:eigen_min} that this is bounded below by $\varepsilon\lambda_0$. The claim thus follows. \qed

%% file: Main_proof.tex
\newpage

\section{Proof of Theorem~\ref{thm:general}}
\label{proof:main}

% \textbf{Additional notation.}

% \begin{definition}[Plausible models]
%     \label{definition:plausible-models}
%     $U\pbrac{X,m}\subseteq\cal{U}$, defined in \eqref{eqn:primitives1} below, is the subset of models for which the mechanism returns message $m\in\cal{M}$ when agent type is $X\in\cal{X}$.
%     \begin{align}
%     U\pbrac{X,m} := \cbrac{u \in \cal{U} : \smap\pbrac{X,u} = m}. \label{eqn:primitives1}
% \end{align}
% \end{definition}

We will use the subscripts ``$0$'' and ``$t$'' with expectations and probabilities for ``$\trueModel\sim\cal{P}_0$,'' and ``$\model_t\sim\trueModel|\cal{F}_{t-1}$'' respectively. Similarly, the subscript ``$0,t$'' will be used for ``$\trueModel\sim\cal{P}_0, \model_t\sim\trueModel|\cal{F}_{t-1}$.''

\subsection{Main proof.}

Fix round $t\in\cbrac{T_0+1, \dots, T}$. For ease of notation, we will drop subscript ``$t$'' from $x_t$ and denote the type of agent $t$ simply by $x$. We will refer to the $\ith{i}$ row of $x$ by $x_i$, and to $\pub(x)$ by $\xpub$.

Our first claim is that if any of the three scenarios listed in the theorem is true, then 
\begin{align}
\pbrac{x_i - x_j} \cdot \pbrac{\model_t - \trueModel} \sim \subG\pbrac{\frac{C^\prime\varepsilon^2\norm{x_i - x_j}_2^2}{\log\pbrac{\frac{2}{\delTS}}}}, \label{eqn:temp-bound}
\end{align}
where $\model_t \sim \trueModel| \cal{F}_{t-1}$, $\trueModel\sim\cal{P}_0$, and $C^\prime>0$ is some universal constant.  Lemmas~\ref{lemma:posterior-confidence-1}, \ref{lemma:posterior-confidence-2} and \ref{lemma:posterior-confidence-3} together prove this claim. Using \ref{eqn:temp-bound} in conjunction with Lemma~ \ref{lemma:subG-tail} then yields 
\begin{align}
&\E_{0,t}\sbrac{\left\lvert \pbrac{x_i - x_j}\cdot \pbrac{\trueModel - \model_t} \right\rvert \indi{ \smap\pbrac{\xpub, \model_t} = m}} \notag \\
\leq\ &\pbrac{\varepsilon/4}\norm{x_i - x_j}_2\sqrt{\frac{\log\pbrac{\frac{2}{\Prob[t]{\smap\pbrac{\xpub, \model_t} = m}}}}{\log\pbrac{\frac{2}{\delTS}}}}\Prob[t]{\smap\pbrac{\xpub, \model_t} = m} \notag \\ 
\underset{\text{Fact~\ref{fact:Posterior-Sampling}}}{=}\ &\pbrac{\varepsilon/4}\norm{x_i - x_j}_2\sqrt{\frac{\log\pbrac{\frac{2}{\Prob[0]{\smap\pbrac{\xpub, \trueModel} = m}}}}{\log\pbrac{\frac{2}{\delTS}}}}\Prob[0]{\smap\pbrac{\xpub, \trueModel} = m} \notag \\
\underset{\eqref{eqn:primitives3}}{\leq}\ &\pbrac{\varepsilon/4}\norm{x_i - x_j}_2\Prob[0]{\smap\pbrac{\xpub, \trueModel} = m}. \label{eqn:temp-bound-2}
\end{align}
Now note that 
\begin{align*}
&\E_{0,t}\sbrac{\pbrac{x_i - x_j}\cdot \trueModel \indi{ M_t = m }} \\
=\ &\E_{0,t}\sbrac{\pbrac{x_i - x_j}\cdot \trueModel \indi{ \smap\pbrac{\xpub, \model_t} = m}} \\
=\ &\E_t\sbrac{\pbrac{x_i - x_j}\cdot \model_t \indi{ \smap\pbrac{\xpub, \model_t} = m}} + \E_{0,t}\sbrac{\pbrac{x_i - x_j}\cdot \pbrac{\trueModel - \model_t} \indi{ \smap(\xpub, \model_t) = m}}\\
\underset{\$_1}{=}\ &\E_0\sbrac{\pbrac{x_i - x_j}\cdot \trueModel \indi{ \smap(\xpub, \trueModel) = m}} + \E_{0,t}\sbrac{\pbrac{x_i - x_j}\cdot \pbrac{\trueModel - \model_t} \indi{ \smap(\xpub, \model_t) = m}}\\
\underset{\dag}{=}\ &\Delta_{i,j}\pbrac{x,m}\Prob[0]{\smap\pbrac{\xpub, \trueModel} = m} + \E_{0,t}\sbrac{\pbrac{x_i - x_j}\cdot \pbrac{\trueModel - \model_t} \indi{ \smap(\xpub, \model_t) = m}}\\
\underset{\star}{\geq}\ &\Delta_{i,j}\pbrac{x,m}\Prob[0]{\smap\pbrac{\xpub, \trueModel} = m} - \pbrac{\varepsilon/4}\norm{x_i - x_j}_2\Prob[0]{\smap\pbrac{\xpub, \trueModel} = m},
\end{align*}
where $(\$_1)$ follows using Fact~\ref{fact:Posterior-Sampling}, $(\dag)$ from \eqref{eqn:primitives2},  and $(\star)$ using \eqref{eqn:temp-bound-2}. The stated assertion in \eqref{eqn:slack} follows after dividing both sides of the inequality in $(\star)$ by $\Prob[0]{\smap\pbrac{\xpub, \trueModel} = m}$ and then using $\Prob[t]{M_t = m} = \Prob[t]{\smap\pbrac{\xpub, \model_t} = m} \underset{\text{Fact~\ref{fact:Posterior-Sampling}}}{=} \Prob[0]{\smap\pbrac{\xpub, \trueModel} = m}$. \qed

\subsection{Auxiliary results used in the main proof}

\begin{lemma}[Confidence bound (i) for posterior sample]
\label{lemma:posterior-confidence-1}

If the warm-up data is non-adaptive, has bandit feedback, and $\emin \geq \lambda_0 >0$ almost surely, then for any fixed $b\in\R^d$, one has $b\cdot\pbrac{\model_t-\trueModel} \sim \subG\pbrac{CR^2C_{\cal{X}}^2\norm{b}_2^2/\lambda_0}$, where $\model_t \sim \trueModel| \cal{F}_{t-1}$, $\trueModel\sim\cal{P}_0$ and $C>0$ is some universal constant.

\end{lemma}

\begin{lemma}[Confidence bound (ii) for posterior sample]
\label{lemma:posterior-confidence-2}

If the warm-up data has bandit feedback, and $\emin \geq \lambda_0 > 0$ almost surely, then for any fixed $b\in\R^d$, one has $b\cdot\pbrac{\model_t-\trueModel} \sim \subG\pbrac{C\pbrac{RC_{\cal{X}} + C_{\cal{U}}}^2d\log\pbrac{C_{\cal{X}}^2t + 3}\norm{b}_2^2/\lambda_0}$, where $\model_t \sim \trueModel| \cal{F}_t$, $\trueModel\sim\cal{P}_0$ and $C>0$ is some universal constant.

\end{lemma}

\begin{lemma}[Confidence bound (iii) for posterior sample]
\label{lemma:posterior-confidence-3}

If the warm-up data has semi-bandit feedback, type space satisfies $\cal{X} \subseteq \cbrac{0,1}^{K\times d}$, and $\eminprime \geq \lambda_0 >0$ almost surely, then for any fixed $\s$-sparse vector $b\in\R^d$, one has $b\cdot\pbrac{\model_t-\trueModel} \sim \subG\pbrac{CR^2\s\norm{b}_2^2/\lambda_0}$, where $\model_t \sim \trueModel| \cal{F}_t$, $\trueModel\sim\cal{P}_0$ and $C>0$ is some universal constant.
\end{lemma}

\begin{lemma}[Tails of sub-Gaussian random variables]
\label{lemma:subG-tail}
If $W\sim\subG\pbrac{A^2}$ and event $E$ satisfies $\Prob{E} \leq \delta$ for some $\delta\in(0,1]$, then $\E\sbrac{\left\lvert W \right\rvert \indi{E} } \leq CA\delta\sqrt{\log\pbrac{2/\delta}}$, where $C$ is some universal constant.

\end{lemma}

\begin{fact}[Standard property of Posterior Sampling]
\label{fact:Posterior-Sampling}
If $\model_t \sim \trueModel | \cal{F}_{t-1}$, then the following holds
    \begin{align}
        &\Prob[t]{\model_t \in \cal{A}} = \Prob[0]{\trueModel \in \cal{A}} \quad \forall\ \text{Borel-measurable subsets $\cal{A}$.} \\
        &\E_{t}\sbrac{f(\model_t)} = \E_{0}\sbrac{f(\trueModel)}\quad \forall\ \text{Borel-measurable functions $f$.}
    \end{align}
\end{fact}

\subsubsection{Proof of Lemma~\ref{lemma:posterior-confidence-1}}

Let $\cal{T} = \sbrac{T_0}$, and consider the static OLS estimator $\hat{\model}_t := \pbrac{\hat{\Sigma}_\cal{T}}^{-1}\sum_{s\in\cal{T}}x_{A_s, s}y_s$.

Note that $b\cdot \pbrac{\hat{\model}_t - \trueModel} = \sum_{s\in\cal{T}} \pbrac{x_{A_s, s} x_{A_s, s}\cdot\pbrac{\hat{\Sigma}_{\cal{T}}}^{-1} b}\cdot {\xi}_s$.  In addition,
\begin{align*}
\sum_{s\in\cal{T}} \norm{x_{A_s, s}x_{A_s, s}\cdot\pbrac{\hat{\Sigma}_{\cal{T}}}^{-1} b}_2^2 = \sum_{s\in\cal{T}} \norm{x_{A_s, s}}_2^2 \pbrac{x_{A_s, s}\cdot\pbrac{\hat{\Sigma}_{\cal{T}}}^{-1} b}^2 &\leq  C_{\cal{X}}^2\sum_{s\in\cal{T}}  \pbrac{x_{A_s, s}\cdot\pbrac{\hat{\Sigma}_{\cal{T}}}^{-1} b}^2 \\
&=  C_{\cal{X}}^2 \norm{b}^2_{\pbrac{\hat{\Sigma}_{\cal{T}}}^{-1}} \\
&\leq \frac{ C_{\cal{X}}^2 \norm{b}_2^2}{\lambda_{\min}\pbrac{ \hat{\Sigma}_{\cal{T}}  }} \\
&\leq \frac{ C_{\cal{X}}^2 \norm{b}_2^2}{\lambda_0}\ \ \ \text{w.p. $1$}.
\end{align*}
Since the action sub-sequence $\pbrac{A_s : s \in\cal{T}}$ is independent of the noise sub-sequence $\pbrac{{\xi}_s : s\in\cal{T}}$, it follows from Assumption~\ref{assumption:noise} and Lemma~\ref{lemma:azuma} that $\left. b\cdot \pbrac{\hat{\model}_t - \trueModel} \right\rvert \trueModel \sim \subG\pbrac{\tilde{R}^2}$, where $\tilde{R}^2 := C^\prime R^2C_{\cal{X}}^2 \norm{b}_2^2/\lambda_0$ for some universal constant $C^\prime>0$.  Therefore,  for any $\varepsilon>0$
\begin{align*}
\Prob{\left.  \left\lvert b\cdot \pbrac{\hat{\model}_t - \trueModel} \right\rvert > \varepsilon \right\rvert \trueModel} \leq 2\exp\pbrac{\frac{-\varepsilon^2}{2\tilde{R}^2}}\ \ \ \forall\ \trueModel\in\cal{U} \underset{\dag}{\implies}\ \Prob[0,t]{\left\lvert b\cdot \pbrac{\model_t - \trueModel} \right\rvert > 2\varepsilon } \leq 4\exp\pbrac{\frac{-\varepsilon^2}{2\tilde{R}^2}},
\end{align*}
where $(\dag)$ follows using Lemma~\ref{lemma:Bayesian-Chernoff}. The claim then follows using Fact~\ref{fact:shamir}.  \qed

\subsubsection{Proof of Lemma~\ref{lemma:posterior-confidence-2}}

Let $\cal{T} = \sbrac{T_0}$ and consider the (static) regularized OLS estimator $\hat{\model}_t := \pbrac{\cal{I}_d + \hat{\Sigma}_{\cal{T}}}^{-1}\sum_{s\in\cal{T}}x_{A_s, s}y_s$. We invoke Lemma~\ref{lemma:OLS-abbasi} with $\delta\gets \delta e$ and obtain the following for any $t\in\N$ and $\delta>0$:
\begin{align}
\Prob{\left. \norm{ \hat{\model}_t - \trueModel  }_{\pbrac{I_d + \hat{\Sigma}_{\cal{T}}}} \geq \pbrac{RC_{\cal{X}} + C_{\cal{U}}}\sqrt{2d\log\pbrac{C_{\cal{X}}^2t + 3}\log\pbrac{\frac{1}{\delta}}}  \right\rvert \trueModel } \leq \delta e\ \ \ \forall\ \trueModel\in\cal{U}. \label{eqn:temp01}
\end{align}
We also know 
\begin{align}
\mod{ b\cdot \pbrac{\hat{\model}_t - \trueModel}} \underset{\text{Fact~\ref{fact:holder}}}{\leq} \norm{b}_{{\pbrac{I_d + \hat{\Sigma}_{\cal{T}}}}^{-1}} \norm{ \hat{\model}_t - \trueModel  }_{\pbrac{I_d + \hat{\Sigma}_{\cal{T}}}} &\underset{\text{Fact~\ref{fact:matrix}}}{\leq} \frac{\norm{b}_2}{\sqrt{\lambda_{\min}\pbrac{I_d + \hat{\Sigma}_{\cal{T}}}}} \norm{ \hat{\model}_t - \trueModel  }_{\pbrac{I_d + \hat{\Sigma}_{\cal{T}}}} \notag \\
&\underset{\text{Fact~\ref{fact:min-eigenvalue}}}{\leq} \frac{\norm{b}_2}{\sqrt{\lambda_{\min}\pbrac{\hat{\Sigma}_{\cal{T}}}}} \norm{ \hat{\model}_t - \trueModel  }_{\pbrac{I_d + \hat{\Sigma}_{\cal{T}}}} \notag \\
&\leq \frac{\norm{b}_2}{\sqrt{\lambda_0}} \norm{ \hat{\model}_t - \trueModel  }_{\pbrac{I_d + \hat{\Sigma}_{\cal{T}}}}\ \ \ \text{w.p. $1$}. \label{eqn:temp02}
\end{align}
From \eqref{eqn:temp01} and \eqref{eqn:temp02}, it follows that
\begin{align*}
    &\Prob{\left. \mod{b\cdot \pbrac{\hat{\model}_t - \trueModel}} \geq \frac{\pbrac{RC_{\cal{X}} + C_{\cal{U}}}\norm{b}_2}{\sqrt{\lambda_0}}\sqrt{2d\log\pbrac{C_{\cal{X}}^2t + 3}\log\pbrac{\frac{1}{\delta}}} \right\rvert \trueModel} \leq \delta e\ \ \ \forall\ \trueModel\in\cal{U} \\
    \underset{\dag}{\implies}\ &\Prob[0,t]{ \mod{b\cdot \pbrac{\model_t - \trueModel}} \geq \frac{2\pbrac{RC_{\cal{X}} + C_{\cal{U}}}\norm{b}_2}{\sqrt{\lambda_0}}\sqrt{2d\log\pbrac{C_{\cal{X}}^2t + 3}\log\pbrac{\frac{1}{\delta}}} } \leq 2\delta e,
\end{align*}
where $(\dag)$ follows using Lemma~\ref{lemma:Bayesian-Chernoff}. The claim now follows using Fact~\ref{fact:shamir}. \qed

\subsubsection{Proof of Lemma~\ref{lemma:posterior-confidence-3}}

Let $T_j(s) := \sum_{u\in[s]}\indi{j\in A_u}$ denote the number of samples of ``atom'' $j\in[d]$ collected up to time $s\in\N$.
Since $\eminprime \geq \lambda_0 >0$ w.p. $1$, and we have semi-bandit feedback,  it follows that at least $N := \left\lceil \lambda_0 \right\rceil$ samples of each atom $j\in[d]$ have been collected by round $T_0$ (w.p.~$1$).  Equivalently,  for each atom $j\in[d]$,  $\exists$ a subset of rounds $\cal{T}_j^N \subseteq \sbrac{T_0}$ s.t.  $\left\lvert \cal{T}_j^N \right\rvert = N$, $j\in A_s\ \forall\ s\in\cal{T}_j^N$, and $\max\pbrac{\cal{T}_j^N} = \inf\cbrac{s\in\N : T_j(s) = N}$. Define $J_{s}^N := \cbrac{j\in [d] : s\in \cal{T}_j^N}$.  Consider the (static) OLS estimator $\hat{\model}_t := \pbrac{ \frac{\sum_{s\in[t]}r_j(s)\indi{j\in J_{s}^N}}{N} : j\in[d]}$.  Observe that 
\begin{align*}
b\cdot \pbrac{ \hat{\model}_t - \trueModel } = \frac{1}{N}\sum_{j\in[d]}\sum_{s\in[t]}b_j\xi_{j,s}\indi{j\in J_{s}^N} = \frac{1}{N}\sum_{j\in[d]}\sum_{s\in\cal{T}_j^N}b_j\xi_{j,s}.
\end{align*}
Using Lemma~\ref{lemma:azuma-sub-sampled},  one concludes that $\sum_{s\in\cal{T}_j^N}b_j\xi_{j,s} \sim \subG\pbrac{C^\prime R^2b_j^2N}$ for some universal constant $C^\prime>0$.  Therefore, it follows from Lemma~\ref{lemma:subG-sum} that $\left. b\cdot \pbrac{\hat{\model}_t - \trueModel} \right\rvert \trueModel \sim \subG\pbrac{C^\prime R^2\norm{b}_1^2/N}$.  Since $N \geq \lambda_0$, and $b$ is $\s$-sparse, we conclude that $\left. b\cdot \pbrac{\hat{\model}_t - \trueModel} \right\rvert \trueModel \sim \subG\pbrac{\tilde{R}^2}$
where $\tilde{R}^2 := C^\prime R^2\s\norm{b}_2^2/\lambda_0$. This implies that for any $\varepsilon>0$, 
\begin{align*}
\Prob{\left.  \left\lvert b\cdot \pbrac{\hat{\model}_t - \trueModel} \right\rvert > \varepsilon \right\rvert \trueModel} \leq 2\exp\pbrac{\frac{-\varepsilon^2}{2\tilde{R}^2}}\ \ \ \forall\ \trueModel\in\cal{U} \underset{\dag}{\implies}\ \Prob[0,t]{\left\lvert b\cdot \pbrac{\model_t - \trueModel} \right\rvert > 2\varepsilon } \leq 4\exp\pbrac{\frac{-\varepsilon^2}{2\tilde{R}^2}},
\end{align*}
where $(\dag)$ follows using Lemma~\ref{lemma:Bayesian-Chernoff}. The stated assertion now follows using Fact~\ref{fact:shamir}.  \qed

\subsection{Proof of Lemma~\ref{lemma:subG-tail}}

Let $\theta^\ast := \sqrt{2A^2\log\pbrac{2/\delta}}$. Note that
\begin{align*}
    \E\sbrac{\mod{W}\indi{E}} \leq \int_{0}^{\infty}\min\pbrac{\delta, \Prob{\mod{W} \geq \theta}} d\theta &\underset{\text{Definition~\ref{definition:subG}}}{\leq} \int_{0}^{\infty}\min\pbrac{\delta, 2\exp\pbrac{\frac{-\theta^2}{2A^2}}} d\theta \\
    &\underset{\text{using $\theta^\ast$}}{\leq} \delta\theta^\ast + 2\int_{\theta^\ast}^{\infty}\exp\pbrac{\frac{-\theta^2}{2A^2}} d\theta \\
    &\underset{\text{Fact~\ref{fact:Gaussian-tail}}}{\leq} \delta\theta^\ast + \frac{2A^2}{\theta^\ast}\exp\pbrac{\frac{-\theta^{\ast 2}}{2A^2}} \\
    &\underset{\text{using $\theta^\ast$}}{=} A\delta\sqrt{2\log\pbrac{2/\delta}}\pbrac{1 + \frac{1}{2\log\pbrac{2/\delta}}} \\
    &\leq \sqrt{2}\pbrac{1 + \frac{1}{2\log 2}} A\delta\sqrt{\log\pbrac{2/\delta}}.
\end{align*} \qed

%% file: OLS_proof.tex
\section{Proof of Theorem~\ref{thm:BIC-OLS}}
\label{proof:BIC-OLS}

\subsection{Main proof}

For ease of notation, we will denote the type of agent $t$ simply by $x$. Fix message $m\in\cal{M}$ and arms $i = \menu_Q(x,m)$ and $j\in\cal{K}\backslash\{i\}$. We will denote the unique element of $\pubtypes$ by $\xpub$. Observe that
\begin{align}
&\E_0\left[ \left( x_{i} - x_j \right)\cdot\trueModel\mathbbm{1}\left\lbrace M_t = m \right\rbrace \right] \notag \\
=\ &\E_0\left[ \left( x_{i} - x_j \right)\cdot\trueModel\mathbbm{1}\left\lbrace \smap\pbrac{\xpub, \model_t} = m \right\rbrace \right] \notag \\
=\ &\E_0\left[ \left( x_{i} - x_j \right)\cdot\trueModel\mathbbm{1}\left\lbrace \smap(\xpub,\trueModel) = m\right\rbrace \right] - \E_0\sbrac{\left( x_{i} - x_j \right)\cdot\trueModel\indi{\smap(\xpub,\trueModel) = m, \smap\pbrac{\xpub,\model_t} \neq m}} \notag \\
{\color{white}\geq}\  &+ \E_0\sbrac{\left( x_{i} - x_j \right)\cdot\trueModel\indi{\smap\pbrac{\xpub,\trueModel} \neq m, \smap\pbrac{\xpub,\model_t} = m}} \notag \\
\geq\ &\E_0\left[ \left( x_{i} - x_j \right)\cdot\trueModel\mathbbm{1}\left\lbrace \smap(\xpub,\trueModel) = m\right\rbrace \right] \notag \\
{\color{white}\geq}\  &- C_{\cal{U}}\norm{x_i - x_j}_2\sbrac{ \bb{P}_0\left( \smap(\xpub,\trueModel) = m, \smap\pbrac{\xpub, \model_t} \neq m\right) + \bb{P}_0\left(\smap\pbrac{\xpub,\trueModel} \neq m, \smap\pbrac{\xpub, \model_t} = m \right)} \notag \\
=\ &\E_0\left[ \left( x_{i} - x_j \right)\cdot\trueModel\mathbbm{1}\left\lbrace \smap(\xpub, \trueModel) = m\right\rbrace \right] - C_{\cal{U}}\norm{x_i - x_j}_2\sbrac{ \bb{P}_0\left( \trueModel \in U_m, \model_t \notin U_m\right) + \bb{P}_0\left(\trueModel \notin U_m, \model_t \in U_m \right)}. \label{eqn:proof-first-reduction-OLS}
\end{align}
We now invoke Lemma~\ref{lemma:posterior-probability-bound} with $\epsilon = \varepsilon/\pbrac{2C_\cal{U}C_{\cal{X}}}$ to bound $\bb{P}_0\left( \trueModel \in U_m, \model_t \notin U_m\right) + \bb{P}_0\left(\trueModel \notin U_m, \model_t \in U_m \right)$. This, combined with \eqref{eqn:proof-first-reduction-OLS}, yields
\begin{align}
    \E_0 \sbrac{ \pbrac{ x_i - x_j }\cdot \trueModel \indi{ M_t = m }} &\geq \E_0 \sbrac{ \pbrac{ x_i - x_j }\cdot \trueModel \indi{ \smap(\xpub, \trueModel) = m } } - \varepsilon\delTS/4 \notag \\
    &\underset{}{\geq} \pbrac{\E_0 \sbrac{ \left. \pbrac{ x_i - x_j }\cdot\trueModel \right\rvert \smap\pbrac{\xpub,\trueModel} = m } - \varepsilon/4 } \Prob[0]{\smap(\xpub,\trueModel) = m}. \label{eqn:proof-second-reduction-OLS}
\end{align}

From $\varepsilon$-menu-consistency of $\smap$, we conclude 
\begin{align}
    \E_0 \sbrac{ \left. \pbrac{ x_{i} - x_j }\cdot\trueModel \right\rvert \smap\pbrac{\xpub,\trueModel} = m }  \underset{}{\geq} \varepsilon, \label{eqn:proof-third-reduction-OLS}
\end{align}
Combining \eqref{eqn:proof-second-reduction-OLS} and \eqref{eqn:proof-third-reduction-OLS}, we get that for $i=\menu_Q(x,m)$ and $j\in \cal{K}\backslash\{i\}$,
\begin{align*}
    &\E_0 \sbrac{ \left. \pbrac{ x_{i} - x_j }\cdot\trueModel \right\rvert M_t = m } \geq \varepsilon\Prob[0]{\smap(\xpub, \trueModel) = m}/4 \underset{}{\geq} \varepsilon\delTS/4.
\end{align*} \qed

\subsection{Auxiliary results used in the main proof}

\begin{lemma}
\label{lemma:posterior-probability-bound}

Suppose Assumptions~\ref{assumption:parameter-set}, \ref{assumption:noise} hold, and $\delTS,\eta$ are positive. 
Fix $\epsilon > 0$,  $t\in\N$, and a point $u_0\in\cal{U}$.  Consider the regularized OLS estimator of $\trueModel$ in round $t$ given by $\model_t := \pbrac{\cal{I}_d + \hat{\Sigma}_{t-1}}^{-1}\sum_{s\in[t-1]}x_{A_s, s}y_s$.  Define $U := \cbrac{u \in \cal{U} : \norm{u - u_0}_{\infty} \leq \epsilon}$. If $\lambda_{\min}\pbrac{\hat{\Sigma}_{[t-1]}} \gtrsim \frac{d^3\pbrac{RC_{\cal{X}} + C_{\cal{U}}}^2}{\epsilon^4\eta^2}\log\pbrac{C_{\cal{X}}^2t + 3}\log\pbrac{\frac{16e}{\epsilon\delTS}}$, then 
\begin{align*}
\bb{P}_0\left(\model_t\in U, \trueModel\notin U \right) + \bb{P}_0\left( \model_t\notin U, \trueModel\in U \right) \leq {\epsilon \delTS}/{4}.
\end{align*}

\end{lemma}

\textit{Proof of Lemma~\ref{lemma:posterior-probability-bound}.}

We will drop the subscript ``0'' in $\E_0$ and $\bb{P}_0$ for brevity of notation. Define the parameter
\begin{align}
\gamma := \frac{\epsilon\eta}{20d}. \label{eqn:gamma-general-general}
\end{align}
Note that $\gamma \in \left( 0, 1 \right)$ since $\epsilon \leqslant 1$, $\eta \leqslant 1$ and $d \geq 1$.  Define $h_d:\left[0,1\right]\mapsto\sbrac{0, 2^d \pbrac{2^d-1}}$ as 
\begin{align}
h_d(\omega) := 2^d\max\pbrac{\left( 1 + \omega \right)^d - 1,\  1 - \pbrac{1-\omega}^d} = 2^d\pbrac{\left( 1 + \omega \right)^d - 1}. \label{eqn:h-general}
\end{align}

Note that ${U}$ can be represented in the following two ways.

\begin{enumerate}

\item ${U} = \tilde{{U}}\cup\partial\tilde{{U}}$ for some $\tilde{{U}},\partial\tilde{{U}}\subseteq\cal{U}$ that satisfy $\tilde{{U}}\cap\partial\tilde{{U}}=\phi$, $\inf_{u\in\tilde{{U}}, v\in\cal{U}\backslash{U}}\norm{u-v}_{2} \geqslant \gamma\epsilon$ and $\bb{P}(\trueModel\in\partial\tilde{{U}}) \leqslant \left({\sup f}\right)\vol{\partial\tilde{{U}}} \leqslant \left({\sup f}\right)h_d(\gamma)\epsilon^d \underset{\dag}\leqslant {\epsilon\delTS}/16$ (where $\dag$ is due to Lemma~\ref{lemma:gamma-bound-general} and \eqref{eqn:primitives-OLS-eta}).

\item ${U} = \bar{{U}}\backslash\partial\bar{{U}}$ for some $\bar{{U}},\partial\bar{{U}}\subseteq\cal{U}$ that satisfy $\inf_{u\in{U}, v\in\cal{U}\backslash\bar{{U}}}\norm{u-v}_{2} \geqslant \gamma\epsilon$ and $\bb{P}(\trueModel\in\partial\bar{{U}}) \leqslant \left({\sup f}\right)\vol{ \partial\bar{{U}}} \leqslant \left({\sup f}\right)h_d(\gamma)\epsilon^d \leqslant {\epsilon\delTS}/16$.\footnote{$\tilde{{U}}$ corresponds to an $\ell_\infty$ ball of radius $(1 - \gamma)\epsilon$ contained inside ${U}$,  while $\bar{{U}}$ to a ball of radius $(1 + \gamma)\epsilon$ that contains ${U}$; both $\tilde{{U}}$ and $\bar{{U}}$ are concentric w.r.t. ${U}$. $\partial\tilde{{U}}$ and $\partial\bar{{U}}$ are the inner and outer shells respectively (w.r.t. ${U}$) of thickness $\gamma\epsilon$ each.}

\end{enumerate}

Now consider
\begin{align}
\bb{P}\left( \trueModel\in{U}, \model_t\notin{U}\right) &\leqslant \bb{P}\left( \trueModel\in\tilde{{U}}, \model_t\notin{U}\right) + \bb{P}\left( \trueModel\in\partial\tilde{{U}} \right) \notag \\
&\leqslant \bb{P}\left( \norm{\model_t - \trueModel}_{2} \geqslant \gamma\epsilon\right) +{\epsilon\delTS}/16 \notag \\
&\underset{Fact~\ref{fact:matrix}}{\leqslant} \bb{P}\left( \norm{\model_t - \trueModel}_{\cal{I}_d + \hat{\Sigma}_{[t-1]}} \geqslant \gamma\epsilon \sqrt{\lambda_{\min}\left( \cal{I}_d + \hat{\Sigma}_{[t-1]} \right)} \right) +{\epsilon\delTS}/16  \notag \\
&\underset{Fact~\ref{fact:min-eigenvalue}}{\leqslant} \bb{P}\left( \norm{\model_t - \trueModel}_{\cal{I}_d + \hat{\Sigma}_{[t-1]}} \geqslant \gamma\epsilon \sqrt{\lambda_{\min}\left( \hat{\Sigma}_{[t-1]} \right)} \right) +{\epsilon\delTS}/16  \notag \\
&\underset{(\star)}{\leqslant} \bb{P}\left( \norm{\model_t - \trueModel}_{\cal{I}_d + \hat{\Sigma}_{[t-1]}} \geqslant \pbrac{RC_{\cal{X}} + C_{\cal{U}}}\sqrt{2d\log\pbrac{C_{\cal{X}}^2t + 3}\log\pbrac{\frac{16e}{\epsilon\delTS}}}\right) +{\epsilon\delTS}/16  \notag \\
&\underset{(\dag)}{\leqslant} {\epsilon\delTS}/16 + {\epsilon\delTS}/16  \notag \\
&= {\epsilon\delTS}/8 , \label{eqn:proof-second-general}
\end{align}
where $(\star)$ holds by assumption on $\lambda_{\min}\left( \hat{\Sigma}_{[t-1]} \right)$,  and $(\dag)$ is due to Lemma~\ref{lemma:OLS-abbasi}.

One can similarly obtain an identical upper bound for $\bb{P}\left(\trueModel\notin{U}, \model_t\in{U} \right)$ as well since
\begin{align}
\bb{P}\left( \trueModel\notin{U}, \model_t\in{U}\right) \leqslant \bb{P}\left( \trueModel\in\cal{U}\backslash\bar{{U}}, \model_t\in{U}\right) + \bb{P}\left( \trueModel\in\partial\bar{{U}}\cap\cal{U} \right) &\leqslant \bb{P}\left( \norm{\model_t - \trueModel}_{2} \geqslant \gamma\epsilon\right) + {\epsilon\delTS}/24  \notag \\
&\leqslant {\epsilon\delTS}/8.  \label{eqn:proof-third-general}
\end{align}

Combining \eqref{eqn:proof-second-general} and \eqref{eqn:proof-third-general} yields the desired result. \qed

\begin{lemma}
\label{lemma:gamma-bound-general}
Consider $\gamma$ as defined in \eqref{eqn:gamma-general-general} and $h_d(\cdot)$ as defined in \eqref{eqn:h-general}. Then, it follows that
\begin{align*}
h_d(\gamma) \leqslant \epsilon\eta{2^d}/16.
\end{align*}
\end{lemma}

\textit{Proof of Lemma~\ref{lemma:gamma-bound-general}.} Note that $h_d^{-1} : \sbrac{0, {2^d}\pbrac{2^d-1}} \mapsto \left[ 0, 1 \right]$ is given by 
\begin{align*}
h_d^{-1}(\tilde{\omega}) := \left( 1 + \frac{\tilde{\omega}}{{2^d}} \right)^{\frac{1}{d}} - 1 ,
\end{align*}
where $d \geqslant 1$. Note that $h_d^{-1}$ is concave and monotone increasing in $\tilde{\omega}$ with $h_d^{-1}(0) = 0$. Therefore, it follows that
\begin{align*}
h_d^{-1}({\tilde{\omega}}) \geqslant \tilde{\omega} \left( h_d^{-1} \right)^\prime(\tilde{\omega}) = \frac{\tilde{\omega}}{d{2^d}\left( 1 +\frac{\tilde{\omega}}{{2^d}} \right)^{1 - \frac{1}{d}}} \geqslant \frac{\tilde{\omega}}{d{2^d}\left( 1 +\frac{\tilde{\omega}}{{2^d}} \right)}.
\end{align*}
Setting $\tilde{\omega}=\epsilon\eta{2^d}/16$, we obtain
\begin{align*}
h_d^{-1}\left( \epsilon\eta{2^d}/16 \right) \geqslant \frac{\epsilon\eta}{16d\left( 1 + \frac{\epsilon\eta}{16} \right)} \geqslant \frac{\epsilon\eta}{20d} \underset{\text{\eqref{eqn:gamma-general-general}}}{=} \gamma.
\end{align*}
The claim now follows by applying $h_d(\cdot)$ to both sides. \hfill $\square$ 

%% file: UCB_proof.tex
\section{Proof of Theorem~\ref{thm:BIC-UCB}}
\label{proof:BIC-UCB}

\subsection{Main proof}

We will drop the subscript ``0'' in $\E_0$ and $\bb{P}_0$ for brevity of notation. Observe that
\begin{align}
\E\sbrac{\pbrac{\trueModel_i - \trueModel_j} \indi{M_t = i}} &\geq \E\sbrac{\pbrac{\trueModel_i - \trueModel_j} \indi{A^\ast = i}} - \sbrac{ \Prob{ A^\ast = i,  M_t \neq i}+ \Prob{A^\ast \neq i, M_t = i}} \notag \\
&\underset{\dag}{\geq} \epsTS\delTS - \sbrac{ \Prob{ A^\ast = i,  M_t \neq i}+ \Prob{A^\ast \neq i, M_t = i}} \notag \\
&\underset{\eqref{eqn:epsUCB}}{=} CK\epsUCB^\alpha - \sbrac{ \Prob{ A^\ast = i,  M_t \neq i}+ \Prob{A^\ast \neq i, M_t = i}}, \label{eqn:proof-first-reduction}
\end{align}
where $\epsTS$ and $\delTS$ in $(\dag)$ are prior-dependent constants defined in \eqref{eqn:primitives-standard}. Note that
\begin{align}
\Prob{ A^\ast = i,  M_t \neq i} = \sum_{j\in[K]\backslash \{i\}}\Prob{ A^\ast = i,  M_t = j} \leq \sum_{j\in[K]\backslash \{i\}}\Prob{ \model^{\texttt{UCB}}_{j,t} \geq \model^{\texttt{UCB}}_{i,t}, \trueModel_i \geq \trueModel_j}. \label{eqn:proof-second-reduction}
\end{align}
Now consider the following
\begin{align}
&\Prob{ \model^{\texttt{UCB}}_{j,t} \geq \model^{\texttt{UCB}}_{i,t}, \trueModel_i \geq \trueModel_j} \notag \\
\leq\ &\Prob{ \model^{\texttt{UCB}}_{j,t} \geq \model^{\texttt{UCB}}_{i,t}, \trueModel_i \geq \trueModel_j + \epsUCB} + \Prob{\trueModel_j \leq \trueModel_i \leq \trueModel_j + \epsUCB} \notag \\
\leq\ &\Prob{\bigcup_{k\in\cbrac{i,j}}\cbrac{\mod{\model^{\texttt{UCB}}_{k,t} - \trueModel_k} \geq \epsUCB/2}} + \Prob{\mod{ \trueModel_i - \trueModel_j } \leq \epsUCB} \notag\\
\leq\ &\Prob{ \mod{\model^{\texttt{UCB}}_{i,t} - \trueModel_i} \geq \epsUCB/2} + \Prob{ \mod{\model^{\texttt{UCB}}_{j,t} - \trueModel_j} \geq \epsUCB/2} + \Prob{\mod{ \trueModel_i - \trueModel_j } \leq \epsUCB} \notag\\
\leq\ &\Prob{ \mod{\model_{i,t} - \trueModel_i} \geq \epsUCB/2 - \sqrt{\frac{\rho\log(t-1)}{N_i(t)}}} + \Prob{ \mod{\model_{j,t} - \trueModel_j} \geq \epsUCB/2 - \sqrt{\frac{\rho\log(t-1)}{N_j(t)}}} \notag \\
{\color{white}\leq}\ &+\Prob{\mod{ \trueModel_i - \trueModel_j } \leq \epsUCB} \notag\\
\underset{\star}{\leq}\ &\Prob{ \mod{\model_{i,t} - \trueModel_i} \geq \epsUCB/4} + \Prob{ \mod{\model_{j,t} - \trueModel_j} \geq \epsUCB/4} + \Prob{\mod{ \trueModel_i - \trueModel_j } \leq \epsUCB} \notag\\
\underset{\dag}{\leq}\ &\frac{C_1\exp\pbrac{-C_2\epsUCB^2 \NUCB } }{\epsUCB^2} + \epsUCB^\alpha, \label{eqn:proof-third-reduction}
\end{align}
where $(\star)$ uses $N_i(t), N_j(t) \gtrsim \rho\log t/\epsUCB^2$ w.p. $1$, and $(\dag)$ is due to Lemma~\ref{lemma:empirical-mean-tail} and Assumption~\ref{assumption:alpha-margin}. Combining \eqref{eqn:proof-second-reduction} and \eqref{eqn:proof-third-reduction}, we conclude that
\begin{align}
    \Prob{ A^\ast = i,  M_t \neq i} \leq \frac{C_1K\exp\pbrac{-C_2\epsUCB^2 \NUCB } }{\epsUCB^2} + K\epsUCB^\alpha. \label{eqn:proof-fourth-reduction}
\end{align}
Proceeding similarly, one can derive an identical upper bound for $\Prob{ A^\ast \neq i,  M_t = i}$ as well, i.e.,
\begin{align}
    \Prob{ A^\ast \neq i,  M_t = i} \leq \frac{C_1K\exp\pbrac{-C_2\epsUCB^2 \NUCB } }{\epsUCB^2} + K\epsUCB^\alpha. \label{eqn:proof-fifth-reduction}
\end{align}
Combining \eqref{eqn:proof-first-reduction}, \eqref{eqn:proof-fourth-reduction} and \eqref{eqn:proof-fifth-reduction}, we get
\begin{align*}
\E\sbrac{\pbrac{\trueModel_i - \trueModel_j} \indi{M_t = i}} \gtrsim K \pbrac{\epsUCB^\alpha - \frac{\exp\pbrac{-C_2\epsUCB^2 \NUCB }  }{\epsUCB^2}}.
\end{align*}
The stated assertion follows after using $\NUCB \gtrsim \pbrac{\frac{\alpha + 2}{\epsUCB^2}}\log\pbrac{\frac{1}{\epsUCB}}$ (from \eqref{eqn:NUCB}) in the inequality above. \qed

\subsection{Auxiliary results used in the main proof}

\begin{lemma}[Deviation bound for empirical mean under adaptive sampling]
\label{lemma:empirical-mean-tail}
Fix some $\epsilon>0$, round $t\in\N$ and arm~$k\in[K]$. If $N_k(t) \geq L$ w.p. $1$, then  $\Prob{\mod{\model_{k,t} - \trueModel_k} \geq \epsilon} \leq \pbrac{C_1/\epsilon^2}\exp\pbrac{-C_2\epsilon^2 L}$ holds with some universal constants $C_1,C_2>0$.
\end{lemma}

\textit{Proof of Lemma~\ref{lemma:empirical-mean-tail}.} Define $\tau_l^k := \inf\cbrac{s\in\N : N_k(s+1) = l}$, i.e., $\tau_l^k$ is the round in which arm~$k$ is played for the $\ith{l}$ time. Note that
\begin{align*}
    \Prob{\mod{\model_{k,t} - \trueModel_k} \geq \epsilon} = \sum_{n = L }^{t} \Prob{\mod{\model_{k,t} - \trueModel_k} \geq \epsilon,\ N_k(t) = n} &\underset{}{=} \sum_{n = L }^{t} \Prob{\mod{ \sum_{l=1}^{n}\xi_{k,\tau^k_{l}} } \geq \epsilon n,\ N_k(t) = n} \\
    &\leq \sum_{n = L }^{t} \Prob{\mod{ \sum_{l=1}^{n}\xi_{k,\tau^k_{l}} } \geq \epsilon n} \\
    &\underset{\text{Lemma~\ref{lemma:azuma-sub-sampled}}}{\leq} \sum_{n = L }^{t} \exp\pbrac{-C_2\epsilon^2 n} \\
    &\leq \frac{\exp\pbrac{-C_2\epsilon^2 L}}{1 - \exp\pbrac{-C_2\epsilon^2} } \\
    &\leq \frac{C_1\exp\pbrac{-C_2\epsilon^2 L}}{\epsilon^2}.
\end{align*} \qed

%% file: corollary_proofs.tex
\section{Proofs of corollaries}

\subsection{Proof of Corollary~\ref{corollary:CB-private}}
\label{proof:corollary-private}

Fix message $m\in\cal{M}$ and type $x\in\cal{X}$, By $\alpha$-menu-consistency of $\smap$, we have $\Delta_{i,j}(x,m) \geq \alpha\ \forall\ j\in\cal{K}$, when $i = \menu_Q(x,m)$. Also, we know that $\norm{x_i - x_j}_2 \leq 2C_{\cal{X}}\ \forall\ i,j\in\cal{K}$. Therefore, using Theorem~\ref{thm:general}, we conclude that \ourAlg is $\pbrac{\alpha-C_{\cal{X}}\varepsilon/2}$-BIC. The stated guarantee can be achieved by substituting $\varepsilon \gets  \varepsilon/C_{\cal{X}}$, effectively increasing $\emin$ by a factor of $C_{\cal{X}}^2$. \qed

\subsection{Proof of Corollary~\ref{corollary:bandit}}
\label{proof:corollary-bandit}

Since the warm-up data contains at least $\NTS$ samples of each arm $i\in[K]$ almost surely, it follows that $\emin \gtrsim \frac{C}{\epsTS^2}\log\pbrac{\frac{2}{\delTS}}$. Also note that for any arm~$j \neq i$, we have $Delta_{i,j}\pbrac{\cal{I}_d,i} \geq \epsTS$. We now invoke Theorem~\ref{thm:general} (scenario (3)) with $\varepsilon = \epsTS$ to conclude the proof. \qed

\subsection{Proof of Corollary~\ref{corollary:sleeping}}
\label{proof:corollary-sleeping}

Since the warm-up data contains at least $\NTS$ samples of each arm $i\in[K]$ almost surely, it follows that $\emin \gtrsim \frac{C}{\epsTS^2}\log\pbrac{\frac{2}{\delTS}}$. Fix message $\scr{R}\in S([K])$, type $x\in\types$, and arm $i = \menu_Q(x,\scr{R})$. Note that for any arm~$j \neq i$, we have  $\Delta_{i,j}\pbrac{x,\scr{R}} \geq \epsTS$. We now invoke Theorem~\ref{thm:general} (scenario (3)) with $\varepsilon = \epsTS$ to conclude the proof. \qed

\subsection{Proof of Corollary~\ref{corollary:combinatorial}}
\label{proof:corollary-combinatorial}

Since the warm-up data contains at least $\NTS$ samples of each atom $j\in[d]$ almost surely, it follows that $\eminprime \gtrsim \frac{C}{\epsTS^2}\log\pbrac{\frac{2}{\delTS}}$. Also note that for any arm~$\armprime \neq \arm$, we have $\Delta_{\arm,\armprime}\pbrac{\xsingleton,\arm} \geq \epsTS$, where $\xsingleton$ denotes the unique element of $\cal{X}$. Note also that $\norm{x_\arm - x_{\armprime}}_2 \leq \sqrt{2\s}\ \forall\ \arm,\armprime\in\cal{K}$. We now invoke Theorem~\ref{thm:general} (scenario (3)) with $\varepsilon = \epsTS/\sqrt{\s}$ to conclude the proof. \qed

\subsection{Proof of Corollary~\ref{corollary:CB-public}}
\label{proof:corollary-public}

Note that the stated condition of spectral diversity implies
\begin{align}
    \emin \gtrsim \frac{R^2}{\pbrac{\alpha/d}^2}\log \pbrac{\frac{2}{\pbrac{\alpha/4}^d}}. \label{eqn:temp-one}
\end{align}
We know from Corollary~3.3 of \cite{sellke2023incentivizing} that under Assumptions~\ref{assumption:alpha-separation} and \ref{assumptipon:uniform-prior-CB},
\begin{align}
    &\Delta_{i,j}\pbrac{x,i} \gtrsim \frac{\alpha\norm{x_i - x_j}_2}{d} \quad \forall\ i,k\in\cal{K},\ x\in\cal{X}\label{eqn:temp-three} \\
    &\delTS \geq \pbrac{\frac{\alpha}{4}}^d. \label{eqn:temp-two}
\end{align}
From \eqref{eqn:temp-one} and \eqref{eqn:temp-two}, we get that $\emin \gtrsim \frac{R^2}{\pbrac{\alpha/d}^2}\log \pbrac{\frac{2}{\delTS}}$. We now use \eqref{eqn:temp-three} together with $\alpha$-separation (Assumption~\ref{assumption:alpha-separation}), and invoke Theorem~\ref{thm:general} with $\varepsilon = \alpha/\pbrac{Cd}$ for some large enough universal constant $C>0$, to complete the proof. \qed

%% file: martingale_stuff.tex
\section{Martingale concentration bounds used in this paper}

\begin{lemma}[Azuma's inequality for martingales with sub-Gaussian tails]
\label{lemma:azuma}
Suppose $\pbrac{\eta_t : t \in \N}$ is an $R$-sub-Gaussian martingale difference sequence w.r.t. filtration $\cbrac{F_t : t  =0,1,...}$, i.e.,  $\eta_t\in F_t$, $\E\sbrac{\eta_t | F_{t-1}} = 0$ and $\eta_t | F_{t-1} \sim \subG\pbrac{R_t^2}\ \forall\ t\in\N$, where $\pbrac{R_t : t\in \N}$ is a real-valued deterministic sequence.  Then, one has that $\sum_{s\in[t]}\eta_s \sim \subG\pbrac{C\sum_{s\in[t]}R_s^2}$,  where $C>0$ is some universal constant.
\end{lemma}

\begin{lemma}[Azuma's inequality for sub-sampled martingales with sub-Gaussian tails]
\label{lemma:azuma-sub-sampled}
Suppose $\pbrac{\eta_t : t \in \N}$ is an $R$-sub-Gaussian martingale difference sequence w.r.t. filtration $\cbrac{F_t : t  =0,1,...}$, i.e.,  $\eta_t\in F_t$, $\E\sbrac{\eta_t | F_{t-1}} = 0$ and $\eta_t | F_{t-1} \sim \subG\pbrac{R^2}\ \forall\ t\in\N$. Fix $N\in\N$ and let $\pbrac{\tau_1, ..., \tau_N}$ be a strictly increasing sequence of $\N$-valued random variables with the property that $\cbrac{\tau_i = t}$ is $F_{t-1}$-measurable $\forall\ t\in\N,\ \forall\ i\in[N]$. Then, one has that $\sum_{i\in[N]}\eta_{\tau_i} \sim \subG\pbrac{CNR^2}$,  where $C>0$ is some universal constant.
\end{lemma}

\begin{lemma}[Sum of sub-Gaussian random variables is sub-Gaussian]
\label{lemma:subG-sum}
Fix $n\in\N$ and $\pbrac{R_i : i\in[n]}\in\R^n$. If $W_i\sim\subG\pbrac{R_i^2}$ for each $i\in[n]$, then $\sum_{i\in[n]}W_i \sim\subG\pbrac{\pbrac{\sum_{i\in[n]}\mod{R_i}}^2}$.
\end{lemma}

\subsection{Proof of Lemma~\ref{lemma:azuma}} 

We know from Markov's inequality that for any $\varepsilon, \theta \geq 0$,
\begin{align*}
\exp\pbrac{\theta\varepsilon}\Prob{\sum_{s\in[t]}\eta_s \geq \varepsilon} \leq \E\sbrac{\exp\pbrac{\theta\sum_{s\in[t]}\eta_s}} &= \E\sbrac{\E\sbrac{\left. \exp\pbrac{\theta\eta_{t}} \right\rvert F_{t-1}}\exp\pbrac{\theta\sum_{s\in[t-1]}\eta_s}} \\ 
&\leq \exp\pbrac{\frac{\theta^2R_t^2}{2}}\E\sbrac{\exp\pbrac{\theta\sum_{s\in[t-1]}\eta_s}} \\
&\leq \dots \\
&\leq \exp\pbrac{\frac{\theta^2\sum_{s\in[t]}R_s^2}{2}}.
\end{align*}
Setting $\theta = \varepsilon/\pbrac{\sum_{s\in[t]}R_s^2}$, we obtain after rearranging $\Prob{\sum_{s\in[t]}\eta_s \geq \varepsilon} \leq \exp \pbrac{\frac{-\varepsilon^2}{2\sum_{s\in[t]}R_s^2}}$. The stated assertion now follows using Fact~\ref{fact:shamir}. \qed 

\subsection{Proof of Lemma~\ref{lemma:azuma-sub-sampled}} 

We know from Markov's inequality that for any $\varepsilon, \theta \geq 0$,
\begin{align*}
\exp\pbrac{\theta\varepsilon}\Prob{\sum_{i\in[N]}\eta_{\tau_i} \geq \varepsilon} \leq \E\sbrac{\exp\pbrac{\theta\sum_{i\in[N]}\eta_{\tau_i}}} &= \E\sbrac{\E\sbrac{\left. \exp\pbrac{\theta\eta_{\tau_N}} \right\rvert F_{\tau_N-1}}\exp\pbrac{\theta\sum_{i\in[N-1]}\eta_{\tau_i}}} \\ 
&\leq \exp\pbrac{\frac{\theta^2R^2}{2}}\E\sbrac{\exp\pbrac{\theta\sum_{i\in[N-1]}\eta_{\tau_i}}} \\
&\leq \dots \\
&\leq \exp\pbrac{\frac{N\theta^2 R^2}{2}}.
\end{align*}
Setting $\theta = \varepsilon/\pbrac{NR^2}$, we obtain after rearranging $\Prob{\sum_{i\in[N]}\eta_{\tau_i} \geq \varepsilon} \leq \exp \pbrac{\frac{-\varepsilon^2}{2N R^2}}$. The stated assertion now follows using Fact~\ref{fact:shamir}. \qed

\subsection{Proof of Lemma~\ref{lemma:subG-sum}}

It suffices to show the result for $n=2$. For any $\theta\in\R$, we have
\begin{align*}
    &\E\sbrac{\exp\pbrac{\theta\pbrac{W_1 + W_2}}} \\
    =\ &\E\sbrac{\exp\pbrac{\theta W_1} \exp\pbrac{\theta W_2}} \\
    \underset{\text{Fact~\ref{fact:holder}}}{\leq}\ &\pbrac{\E\sbrac{\exp\pbrac{\pbrac{\frac{\mod{R_1} + \mod{R_2}}{\mod{R_1}}}\theta W_1} }}^\pbrac{\frac{\mod{R_1}}{\mod{R_1} + \mod{R_2}}}  \pbrac{\E\sbrac{\exp \pbrac{\pbrac{\frac{\mod{R_1} + \mod{R_2}}{\mod{R_2}}} \theta W_2} }}^\pbrac{\frac{\mod{R_2}}{\mod{R_1} + \mod{R_2}}} \\
    \underset{\text{Definition~\ref{definition:subG}}}{=}\ &\exp\pbrac{\frac{\theta^2\pbrac{\mod{R_1} + \mod{R_2}}^2}{2}}.
\end{align*} \qed